\documentclass[11pt]{article}
\usepackage{samaya}
\usepackage{lipsum}   %
\usepackage[numbers]{natbib}  %

\usepackage{subcaption}
\usepackage{graphicx}

\usepackage{listings}
\usepackage{xcolor}   %

\let\samayathead\thead
\let\thead\relax
\usepackage{booktabs,tabularx,array,makecell,xcolor,colortbl,pifont,caption}
\let\thead\samayathead

\lstdefinestyle{judgeprompt}{
  basicstyle=\ttfamily\footnotesize,
  breaklines=true,
  breakindent=0pt,
  columns=fullflexible,
  keepspaces=true,
  frame=single,
  framesep=6pt,
  xleftmargin=6pt,
  xrightmargin=6pt,
}

\usepackage[colorinlistoftodos,textsize=small]{todonotes}

\newcommand{\frontierfinance}{FrontierFinance}

\title{\frontierfinance: A Challenging Benchmark for Measuring Frontier Intelligence of Finance Agents}

\author{Yuhao Zhang, O. Ozan Koyluoglu, Thejas Venkatesh, Richard Diehl Martinez, Vishank Bhatia,  Arash Alidoust, Ashwin Paranjape\\[4pt]
\samayaaffil{Samaya AI}
}

\samayacorrespondence{\samayaemail{research@samaya.ai}}

\begin{document}

\begin{samayatitle}
AI agents are increasingly deployed for professional investment research, yet no benchmark captures the complexity of the full investor workflow.
Existing benchmarks mainly target financial data extraction---a narrow slice that current models have largely saturated---while reference-based metrics and generic LLM-as-a-judge scoring fall short on the open-ended, long-form answers that real analyst queries demand.
We introduce \frontierfinance, a fully open benchmark of 220 expert-crafted queries and 11{,}543 source-attributed rubrics spanning six crucial use cases across the full investor workflow.
\frontierfinance{} is both broader and harder than existing public finance benchmarks.
Evaluating frontier models and agent systems under a common harness restricted to publicly available data, we find that the tool harness, not the model alone, strongly shapes quality and efficiency;
that Samaya's in-house system leads at 56.0\%, ahead of the strongest frontier model (Claude Fable 5, 49.2\%) at roughly 2.2$\times$ lower cost; and that the best open-weight model (Kimi K3, 46.4\%) nearly matches the best proprietary model at 4.5$\times$ lower cost.
\emph{Screening \& Discovery} and \emph{Sector, Industry \& Macro} remain the hardest use cases across all systems, where even the best systems reach only 33\% and 39\%.
We make the dataset and grading code publicly available.

\samayawebsite{%
    \samayaurl
      {https://research.samaya.ai/benchmarks/frontier-finance}
      {https://research.samaya.ai/benchmarks/frontier-finance}%
}

\end{samayatitle}

\section{Introduction}
Large language models are becoming increasingly capable at agentic tasks: they create multi-step plans, call external tools, retrieve evidence from large corpora, and produce long-form outputs combining structured and unstructured content.
Finance is among the most demanding settings for such systems. Professional investment research is open-ended and challenging: it spans idea screening and discovery, company and market research, financial data extraction, portfolio tracking, and catalyst monitoring.
Answering a real-world financial query often requires retrieving and synthesizing qualitative and quantitative evidence from many sources, performing causal analysis, and drafting a comprehensive report that meets a strict standard of factual accuracy.

Existing benchmarks capture only a fraction of this work (Table~\ref{tab:benchmark_landscape}). Most public finance evaluation suites focus primarily on financial data extraction, e.g., retrieving a reported figure from a filing or computing a ratio from a table \citep{islam2023financebench, chen2021finqa, zhu2021tatqa, chen2022convfinqa}.
These tasks are well-defined but narrow, and recent models have largely saturated them. Performance on these static benchmarks is further confounded by data memorization, where models rely on memorized training data rather than true reasoning.
Meanwhile, benchmarks probing complex, open-ended financial research remain scarce. The field therefore lacks a shared and sufficiently difficult benchmark for real-world finance agent evaluation.

To fill this gap, we introduce \frontierfinance, a publicly released benchmark for evaluating finance agents on professional investment research.
It consists of 220 queries and 11,543 rubrics authored through a four-stage curation and audit pipeline by finance domain experts---including former buy-side analysts, sell-side research associates, and investment banking professionals---organized into six use cases spanning the investor workflow (Section~\ref{sec:collection}).
Each rubric is a binary criterion attributed to a publicly available data source, enabling objective, reproducible scoring via a rubric qualification rate---the majority verdict of three independent LLM judges (Section~\ref{sec:experiments}).
Compared to existing benchmarks, \frontierfinance{} is more diverse in use case coverage and significantly harder, and to our knowledge is the largest of its kind (Section~\ref{sec:difficulty}).

We benchmark frontier models and agent systems under a common harness restricted to publicly available data, measuring quality, cost, and latency (Section~\ref{sec:results}).
We find that the harness system, not the underlying model alone, strongly shapes both quality and efficiency. Samaya's in-house system leads at 56\%, ahead of the strongest frontier model deployed under an open-source harness (Claude Fable 5, 49.2\%) at roughly 2.2$\times$ lower cost.
The best open-weight model (Kimi K3, 46.4\%) nearly matches frontier performance at 4.5$\times$ lower cost.
Across all tested systems, we find two use cases, both open-ended in nature, prove hardest and remain largely unsolved: \emph{Screening \& Discovery} and \emph{Sector, Industry \& Macro}, where the best systems reach only 33\% and 39\%.

Beyond aggregate scores, we analyze agent trajectories to understand how systems differ in behavior (Section~\ref{sec:trajectory-analysis}).
Our analysis reveals that tool use follows a common three-phase structure across all systems: data gathering, mid-rollout synthesis and analysis, and answer preparation.
Despite very different tool-call volumes, top systems tend to display more efficient tool use and converge on similar token budgets.
We also uncover failure patterns that may inspire future research: models that navigate directly to known financial sources from parametric knowledge---rather than discovering them through search---incur significantly higher URL error rates, causing token waste and context pollution.

To summarize, our contributions are as follows:
\begin{itemize}
\item We release \frontierfinance, an open benchmark of 220 expert-crafted queries and 11{,}543 source-attributed rubrics spanning six use cases across the full investor workflow, the largest open finance agent benchmark of its kind.
\item We characterize the benchmark's coverage and difficulty through detailed analysis, showing it is both broader and significantly harder than existing public finance benchmarks.
\item We systematically evaluate frontier models and agent systems along the quality, cost, and latency dimensions, and report findings on the role of the tool harness, the quality--cost frontier, and the competitiveness of open-weight models.
\item Through agent trajectory analysis, we identify common behavior patterns as well as pitfalls that point to opportunities for improving future models.
\end{itemize}

\section{Related Work}

\textbf{Early financial QA benchmarks.}\quad Financial benchmarks have evolved from document-grounded question answering toward open-ended, agentic research tasks. Early datasets primarily evaluate numerical reasoning over bounded financial documents: FinQA represents solutions as executable programs, TAT-QA combines tabular and textual evidence, and ConvFinQA extends numerical reasoning to multi-turn conversations \citep{chen2021finqa,zhu2021tatqa,chen2022convfinqa}. FinanceBench evaluates question answering over public-company filings, while DocFinQA and FinanceReasoning introduce longer contexts and more challenging numerical derivations \citep{islam2023financebench,reddy2024docfinqa,tang2025financereasoning}. Broader suites such as FinBen aggregate multiple financial-language tasks into a common evaluation \citep{xie2024finben}. These benchmarks provide controlled tests of component capabilities, but most supply a bounded evidence set and expect a compact answer, numerical value, or derivation. Consequently, none measures whether an agent can discover relevant evidence, scope the research, and synthesize a comprehensive response to an open-ended analyst request.

\definecolor{FFBrand}{HTML}{0019FF}      %
\definecolor{FFBrandInk}{HTML}{0012B2}   %
\definecolor{FFBrandTint}{HTML}{F0F5FF}  %
\definecolor{FFInk}{HTML}{181A1E}        %
\definecolor{FFInkSoft}{HTML}{3E4349}    %
\definecolor{FFInkMuted}{HTML}{777E85}   %
\definecolor{FFRule}{HTML}{D7DDD9}       %
\definecolor{FFRuleSoft}{HTML}{EEEFEA}   %
\definecolor{FFUp}{HTML}{299448}         %
\definecolor{FFDown}{HTML}{E01F1A}       %

\newcommand{\FFbody}{\fontsize{8.5}{10.2}\selectfont}
\newcommand{\FFmeta}{\fontsize{7}{8.2}\selectfont}

\newcommand{\FFyes}{\textcolor{FFInk}{\fontsize{9}{9}\selectfont$\bullet$}}
\newcommand{\FFno}{\textcolor{FFInkMuted}{\fontsize{10.5}{10.5}\selectfont$\circ$}}
\newcommand{\FFna}{\textcolor{FFInkMuted}{--}}

\newcommand{\FFhead}[1]{{\FFbody\bfseries\makecell[bc]{#1}}}
\newcommand{\FFheadleft}[1]{{\FFbody\bfseries\makecell[bl]{#1}}}
\newcommand{\FFbench}[2]{%
  {\FFbody\bfseries #1}~%
  {\FFbody\color{FFInkMuted}%
   \begingroup\hypersetup{citecolor=FFInkMuted,linkcolor=FFInkMuted}%
   \citeyearpar{#2}\endgroup}}
\newcommand{\FFgroup}[1]{%
  \rule[-0.7ex]{0pt}{2.75ex}%
  \FFmeta\bfseries\scshape\color{FFInkSoft}#1}

\newcolumntype{L}[1]{>{\raggedright\arraybackslash}m{#1}}
\newcolumntype{C}[1]{>{\centering\arraybackslash}m{#1}}
\renewcommand{\tabularxcolumn}[1]{m{#1}}

\begin{table}[!t]
\centering
\captionsetup{
  font=small,
  labelfont=bf,
  position=t,
  justification=raggedright,
  singlelinecheck=false,
  skip=7pt
}
\caption[Comparison of finance-agent benchmarks.]{%
\textbf{Comparison of representative finance-agent benchmarks.}
Rows are grouped by their primary evaluation target.
The first five columns indicate whether each benchmark centrally evaluates:
\emph{open-domain discovery}, in which the agent must locate evidence from open search rather
than receive a fixed document corpus;
\emph{long-form output}, in which the principal deliverable is a structured
research report;
\emph{workflow breadth}, meaning coverage of multiple stages of the
professional investor workflow rather than a single task family;
\emph{expert criteria}, meaning task-specific grading requirements authored
by finance-domain experts; and
\emph{source labels}, meaning criterion-level annotation of the expected
evidence category.
The final column reports the number of publicly released examples.}
\label{tab:benchmark_landscape}

\FFbody
\setlength{\tabcolsep}{2.9pt}
\renewcommand{\arraystretch}{1.15}
\begin{tabularx}{\textwidth}{
  @{}
  L{3.2cm}
  >{\raggedright\arraybackslash}X
  *{5}{C{1.15cm}}
  C{1.5cm}
  @{}
}
\arrayrulecolor{FFInk}\toprule
\multicolumn{1}{@{}l}{\FFheadleft{Benchmark}} &
\FFheadleft{Primary emphasis} &
\FFhead{Open\\domain} &
\FFhead{Long-\\form} &
\FFhead{Workflow\\breadth} &
\FFhead{Expert\\criteria} &
\FFhead{Source\\labels} &
\FFhead{Public\\examples} \\
\arrayrulecolor{FFInk}\midrule

\rowcolor{FFRuleSoft}
\multicolumn{8}{@{}l}{\FFgroup{Document-grounded question answering}} \\[2.5pt]

\FFbench{FinanceBench}{islam2023financebench} &
Filing-grounded financial QA &
\FFno & \FFno & \FFno & \FFno & \FFno & 150 \\
\addlinespace[3pt]

\rowcolor{FFRuleSoft}
\multicolumn{8}{@{}l}{\FFgroup{Agentic retrieval and derivation}} \\[2.5pt]

\FFbench{Finance Agent Benchmark}{bigeard2025financeagentbenchmark} &
SEC filings and web research &
\FFyes & \FFyes & \FFno & \FFyes & \FFno & 27 \\

\FFbench{BigFinance\allowbreak{}Bench}{wang2026bigfinancebench} &
Auditable financial derivations &
\FFyes & \FFno & \FFno & \FFyes & \FFno & 50 \\

\FFbench{Hedge-Bench}{cho2026hedgebench} &
Expert steps in a fixed corpus &
\FFno & \FFno & \FFno & \FFyes & \FFno & 102 \\
\addlinespace[3pt]

\rowcolor{FFRuleSoft}
\multicolumn{8}{@{}l}{\FFgroup{Open-ended research reports}} \\[2.5pt]

\FFbench{FinResearch\allowbreak{}Bench II}{luan2026finresearchbenchii} &
Rubric-scored research reports &
\FFyes & \FFyes & \FFno & \FFno & \FFno & 0 \\

\arrayrulecolor{FFInk}\midrule

\rowcolor{FFBrandTint}
{\FFbody\bfseries\color{FFBrandInk}Frontier\allowbreak{}Finance} &
{\FFbody\bfseries Full investor research workflow} &
\FFyes & \FFyes & \FFyes & \FFyes & \FFyes &
{\FFbody\bfseries 220} \\

\arrayrulecolor{FFInk}\bottomrule
\end{tabularx}
\end{table}

\textbf{Agentic finance benchmarks.}\quad More recent benchmarks explicitly evaluate financial agents equipped with search and specialized tools. The Finance Agent Benchmark contains expert-authored research problems requiring agents to use recent filings and web evidence, while FinSearchComp evaluates time-sensitive retrieval, historical lookup, and complex open-domain financial investigation \citep{bigeard2025financeagentbenchmark,hu2025finsearchcomp}. FinAgentBench and Fin-RATE focus on agentic retrieval and longitudinal or cross-company reasoning over regulatory filings \citep{choi2025finagentbench,jiang2026finrate}. Other benchmarks broaden the domain beyond filing research: FinGAIA evaluates multi-tool financial agents, FORCE-Bench targets operational and enterprise-finance workflows, Herculean covers various financial-intelligence tasks, and FinanceComplexQA evaluates complex reasoning over industrial financial documents \citep{zeng2025fingaia,pauli2026forcebench,peng2026herculean,cheng2026financecomplexqa}. Together, these works extend evaluation to include retrieval systems, tools, and agent orchestration.

\textbf{Evaluating open-ended research answers.}\quad Evaluating open-ended, long-form research answers challenges standard methodology. Open-ended analyst requests admit multiple valid choices of evidence, organization, peer set, and analytical path, making exact match and lexical reference metrics insufficient \citep{xu2023criticalevaluation}. Holistic LLM judges offer greater flexibility and can align with human preferences, but their scores may be sensitive to the judge model, prompt, presentation order, response style, and verbosity, while revealing little about what a system actually missed \citep{liu2023geval,zheng2023llmasajudge}. Fine-grained evaluation avoids these pitfalls by decomposing quality into independently assessable requirements. Such checklist-based methods improve interpretability and inter-judge reliability, and recent long-form research benchmarks increasingly employ task-specific criteria rather than a single overall rating \citep{arora2025healthbench,lee2025checkeval,sharma2025researchrubrics,ruan2025expertlongbench}. In earlier work, we proposed Criteria-Eval \citep{baziotis2025criteriaeval}, a checklist-based framework in which finance experts author binary criteria and score systems by the fraction satisfied---an approach that jointly handles retrieval and generation, admits multiple valid answers, and aligns scoring with expert judgment.

\textbf{Open-ended financial research benchmarks.}\quad The closest precedents to our setting evaluate open-ended financial research or professional work products. BigFinanceBench decomposes financial research into point-weighted, independently checkable derivation steps, while FinResearchBench evaluates reports through an intermediate logic tree and FinResearchBench II constructs query-specific rubrics from model-generated reports \citep{wang2026bigfinancebench,sun2025finresearchbench,luan2026finresearchbenchii}. FinDeepResearch evaluates standardized company-analysis reports across markets and languages, Deep FinResearch Bench compares generated research with professional analyst reports, and Hedge-Bench scores verified expert reasoning steps within controlled evidence environments \citep{zhu2025findeepresearch,haque2026deepfinresearchbench,cho2026hedgebench}. Complementary artifact-oriented benchmarks focus on constructing financial spreadsheets and models rather than research reports \citep{yen2026mbabench,krumdick2026frontierfinance}.

\section{Data Collection and Statistics}

\subsection{Data collection}\label{sec:collection}
We build \frontierfinance{} through a four-stage curation and audit pipeline executed entirely by finance experts---including former buy-side equity analysts, sell-side research associates, and investment banking professionals. Figure~\ref{fig:data-collection-diagram} illustrates this pipeline.
We include examples of the resulting annotations in Appendix~\ref{appendix:samples}.

\begin{figure}[htbp]
  \centering
  \includegraphics[width=0.99\textwidth]{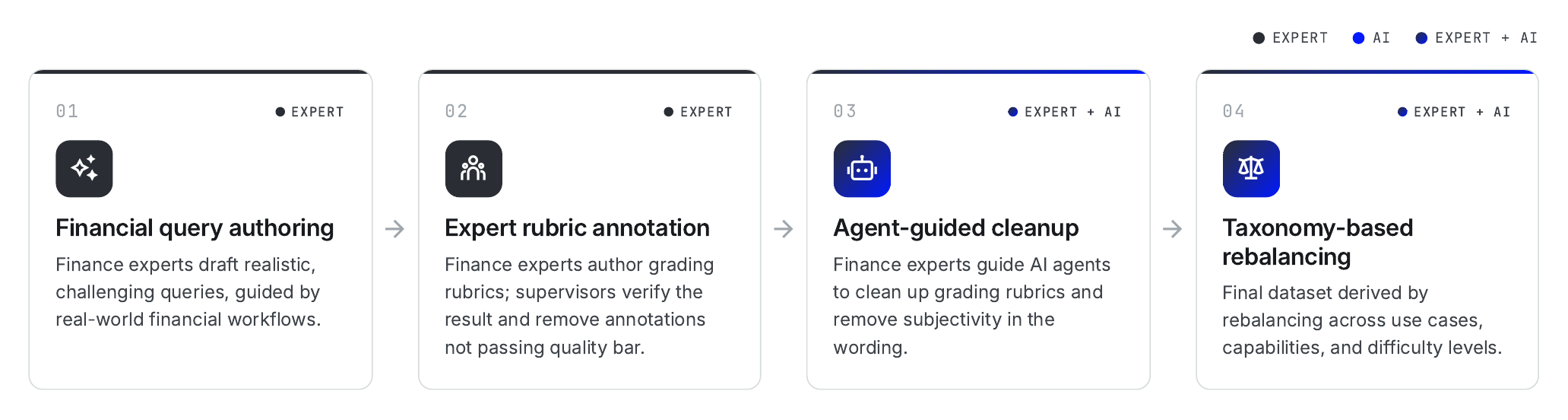}
  \caption{The four-stage data collection process used to construct \frontierfinance{}, spanning query drafting, dense rubric authoring, multi-expert auditing, and dataset rebalancing.}
  \label{fig:data-collection-diagram}
\end{figure}

\begin{figure}[t]
\centering
\includegraphics[width=\linewidth]{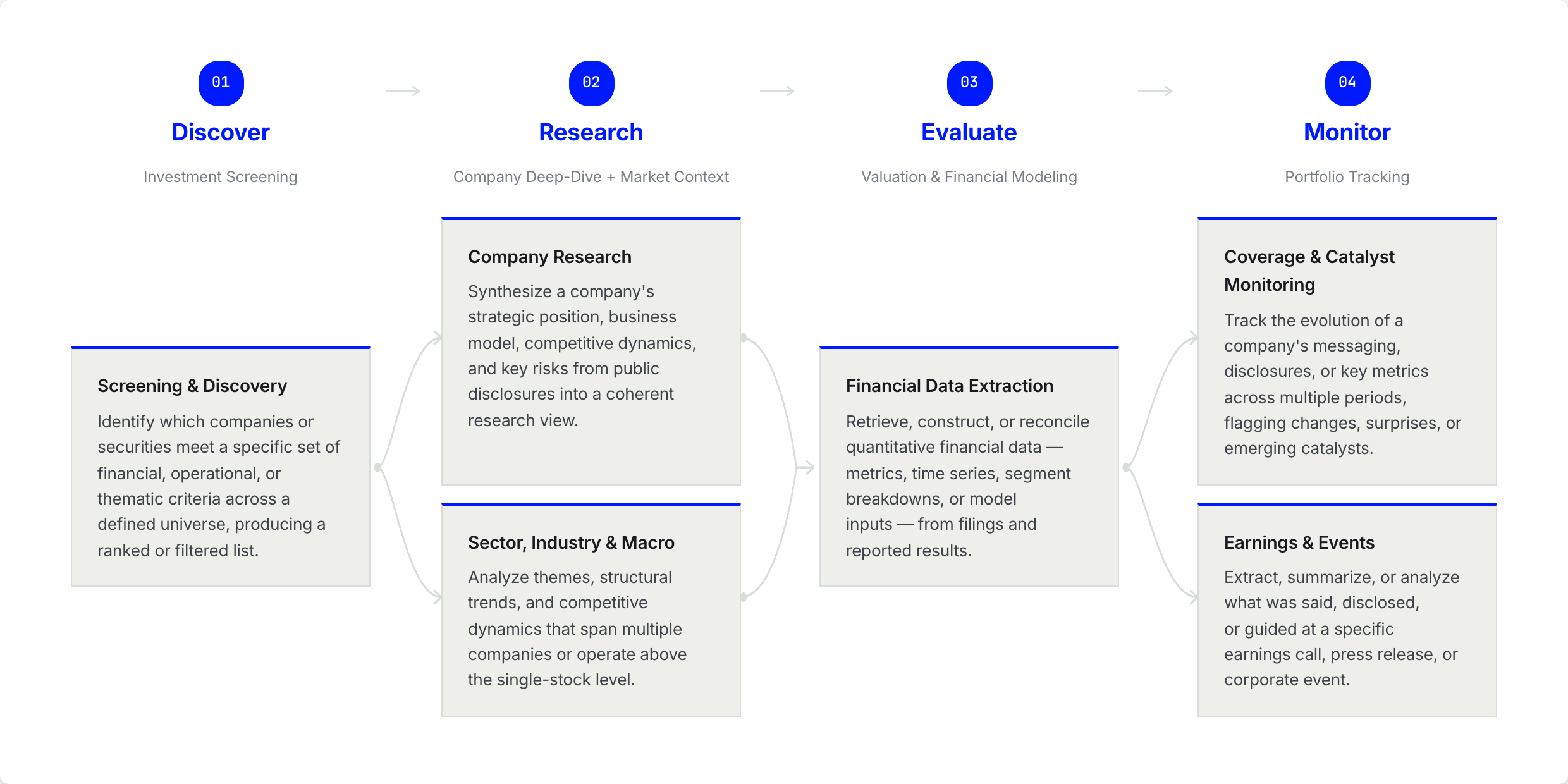}
\vspace{-2em}
\caption{FrontierFinance focuses on 6 use cases that span the entire investment decision-making workflow.}
\label{fig:usecases}
\end{figure}

\textbf{Stage 1: Workflow mapping and query drafting.}
Domain experts first mapped the end-to-end investment workflow, identified six use cases (Figure~\ref{fig:usecases}), and simulated investment decision-making to craft open-ended queries reflecting real analyst tasks. To test agent robustness, queries deliberately preserve real-world ambiguity: using company names and ticker symbols interchangeably without explicit mapping, specifying multi-quarter or dynamic timeframes that require resolving historical date bounds against filing dates, and framing tasks requiring unbounded retrieval across market caps and peer groups rather than closed context windows.

\textbf{Stage 2: Rubric authoring and source attribution.}
For each query, experts constructed binary evaluation rubrics that decompose open-ended research deliverables into independently checkable criteria. Each rubric is annotated with \emph{source attribution}, tying it to a primary public data source tier such as SEC filings, call transcripts, or market data (Section~\ref{sec:sources}). Queries average 52.5 rubrics each, enabling thorough grading of long-form answers.

\textbf{Stage 3: Expert review and multi-stage auditing.}
Under the help of AI agents, a first panel of experts reviews drafted queries and rubric sets to eliminate subjective phrasing, repeated information, and language that requires a single source when multiple are possible. A secondary panel verifies that every criterion can be objectively evaluated as fulfilled (\texttt{1}) or unfulfilled (\texttt{0}) against public evidence, consolidates redundant rubrics to prevent over-weighting repeated facts, and confirms that all attributed evidence is accessible via public web or regulatory channels without proprietary paywalls. We also tag each rubric along two dimensions: \emph{essentiality} (mandatory \emph{must-have} vs.\ \emph{supplementary}) and \emph{rubric category} (an eight-category functional taxonomy; Section~\ref{sec:composition}).

\textbf{Stage 4: Rebalancing and dataset stratification.}
Steps 1 through 3 yielded over $4{,}500$ fully annotated research queries. From this pool, we assembled the 220 released queries ($11{,}543$ total rubrics) comprising \frontierfinance{} via stratified sampling, reserving the remaining $\sim\!4{,}300$ queries for internal development and future releases. The public set is balanced across three axes: \emph{use cases} (ensuring sufficient coverage across all six), \emph{capabilities} (stratifying across reasoning and search modalities like temporal filtering and cross-entity triangulation), and \emph{difficulty} (calibrating across Bradley--Terry difficulty terciles; Section~\ref{sec:difficulty}).

\textbf{Timestamped annotation.}\label{sec:timestamped-annotation}
One design decision runs through every stage of this pipeline: how we handle time. Time is central to finance, shaping both how a query should be interpreted (a request for the ``latest quarter'' resolves to different periods depending on when it is asked) and what constitutes a good answer (prices, filings, and consensus estimates change continuously). Handling this time-variant nature of real-world questions is a long-standing challenge, and existing benchmarks address it in one of three ways: restricting to time-invariant queries \cite{wei2025browsecomp, mialon2023gaiabenchmark}, which sharply limits query diversity and is especially restrictive for finance tasks; freezing the data source to a point-in-time corpus \cite{chen2025browsecompplus, du2025deepresearchbench}, which is much smaller and less diverse than the open web and often does not reflect how an agent is deployed in practice; or annotating timestamped queries and rubrics \cite{bigeard2025financeagentbenchmark}, where each query carries a date and the agent answers as if at that date.

We follow this third approach: every \frontierfinance{} query carries a date field, and its rubrics reflect the state of the world up to that date. We further exclude predictive queries (for example, forecasting the outcome of a future event), which keeps the benchmark robust to the exact web snapshot an agent sees even when data postdating the query date is accessible.

\subsection{Use cases, capabilities, and rubric categories}\label{sec:composition}

Examples in \frontierfinance{} are further tagged along three complementary axes: each query carries a
single \textbf{use case} and one or more \textbf{capabilities}, and each rubric is
assigned a \textbf{rubric category}. Together they characterize what the benchmark
asks of an agent and what a correct answer must contain.

\begin{figure}[t]
\centering
\includegraphics[width=0.9\linewidth]{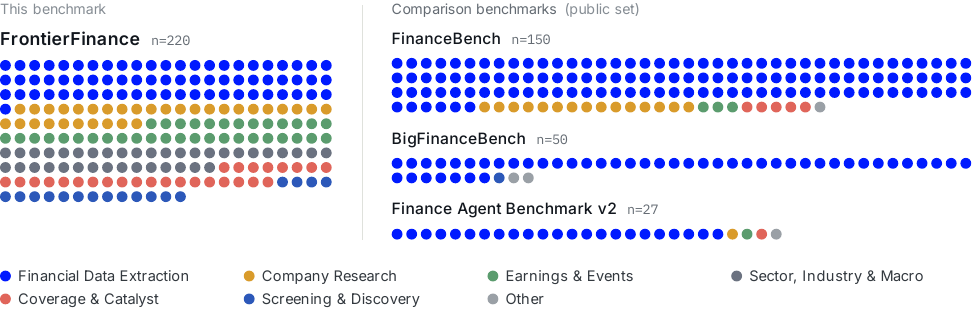}
\caption{Use-case coverage of FrontierFinance versus public finance benchmarks. Each dot is one query, colored by its use case (grey = outside the six use cases). FrontierFinance spreads its 220 queries across all six use cases, whereas existing benchmarks concentrate almost entirely in financial-data extraction.}\label{fig:usecases_dist}
\end{figure}

\textbf{Use cases.}\quad Each query is labelled with exactly one of six use cases
spanning the investor workflow (Figure~\ref{fig:usecases}; full counts and
descriptions in Appendix~\ref{sec:appendix_datastats},
Table~\ref{tab:usecases}). Financial data extraction is the most common (32\% of queries), but no use case dominates, and the
open-ended use cases---screening and discovery, and sector, industry, and
macro---are well represented.
In comparison, existing finance benchmarks concentrate almost entirely on financial-data-extraction queries, leaving harder, open-ended use cases largely untested (Figure~\ref{fig:usecases_dist}).

\textbf{Capabilities.}\quad Each query is also tagged with the reasoning and
retrieval capabilities it requires. Queries average 1.9 capabilities each (410 tags
over 220 queries), reflecting that realistic tasks combine skills. Qualitative
synthesis and the three modes of exhaustive retrieval (temporal, cross-entity, and
thematic; 147 queries combined) are the most common
(see Appendix~\ref{sec:appendix_datastats}, Table~\ref{tab:capabilities} for full counts and definitions).

\textbf{Rubric categories.}\quad Each of the 11,543 rubrics is classified by content type using an eight-category taxonomy
(Table~\ref{tab:rubriccats}, in Appendix~\ref{sec:appendix_datastats}). Factual data
extraction is the plurality (74\%), consistent with rubrics being written as
objective, checkable criteria; the remaining quarter covers qualitative, forward-looking, analytical, and comparative content.

\section{Data Analysis}\label{sec:data-analysis}

We analyze the collected queries and rubrics in depth to characterize their data-source demands and difficulty relative to existing benchmarks.

\subsection{Data sources distribution}\label{sec:sources}

We assign a data source to each rubric using a taxonomy of 10 top-level source categories. Attributing evidence at the rubric level, rather than the query level, lets us characterize the evidence demands of individual queries, use cases, or the benchmark as a whole. The resulting distribution (Appendix~\ref{sec:appendix_datastats}) shows that no single document type dominates.

\textbf{\frontierfinance{} covers diverse data sources demanded by professional finance work.}
SEC filings are the single largest source category, yet they account for under 40\% of all rubrics. The remainder spans company-issued material (earnings-call transcripts, investor presentations, earnings releases, annual reports), the analyst's own professional knowledge (valuation work, model estimates, domain synthesis), live market data, news and media, and regulatory filings.

\begin{figure}[t]
\centering
\includegraphics[width=0.9\linewidth]{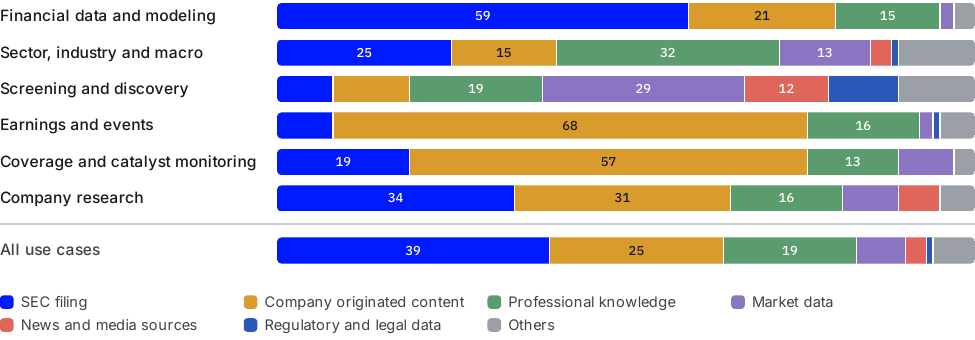}
\caption{Rubric-level data source distribution per use case. Each bar shows how a use case's rubrics break down across data sources (row-normalized to 100\%); the bottom bar is all use cases combined. SEC filings are the plurality overall (39\%) yet under 40\% of rubrics, and the distribution shifts sharply by use case---from filings-driven financial data extraction to company-issued content for earnings and market/professional sources for screening. Full distribution data shown in Appendix \ref{sec:appendix_datastats}, Table~\ref{tab:rubric-data-source}.}
\label{fig:sourcexuse}
\end{figure}

\textbf{The source mix shifts substantially across use cases} (Figure~\ref{fig:sourcexuse}), mirroring how an analyst's evidence needs change
across the workflow. Financial data extraction is overwhelmingly filings-driven
(59\% SEC); earnings \& events leans on company-issued content (68\%, transcripts
and releases); sector/industry \& macro is carried by professional knowledge (32\%)
and market data (13\%); and screening \& discovery is the most source-diverse use
case, pulling from market data (29\%), professional knowledge (20\%), regulatory
data (10\%), and news (12\%) rather than any single dominant source.

\subsection{Difficulty analysis}\label{sec:difficulty}\label{sec:comparison}

We define \emph{difficulty} as the data-gathering and reasoning effort a financial analyst would need to produce a complete, defensible answer. To place every query on a single scale, we use pairwise judgments aggregated via a Bradley--Terry (BT) model---a well-established approach used by Chatbot Arena to rank models \citep{zheng2023llmasajudge, chiang2024chatbot} and underlying alignment methods such as DPO \citep{rafailov2023direct}. Unlike those model-ranking uses, here we rank queries by difficulty. We score each pair using a consensus of three independent LLM judges reasoning over five axes: \textbf{retrieval breadth}, \textbf{reasoning depth}, \textbf{entity scope}, \textbf{time scope}, and \textbf{qualitative ambiguity}. From $\sim$77K pairwise judgments across 4K internal queries, we fit a confidence-weighted BT model \citep{bradley1952rank} (Appendix~\ref{sec:appendix_difficulty}), yielding one latent difficulty score per query.

\begin{table}[t]
\centering\small
\caption{Difficulty terciles of the \frontierfinance{} public set (220 queries), by consensus Bradley-Terry (BT) score (higher means more difficult). On all \frontierfinance{} queries, we see a mean score of 4.84 and standard deviation of 7.16. The bucket cutoffs are easy/medium at $\mathrm{BT} = +0.94$ and medium/hard at $\mathrm{BT} = +7.25$. In the last column, we also include Samaya system's performance for queries in each difficulty bucket as a reference (using must-have qualification rate as defined in experiments).}
\label{tab:difficulty-terciles}
\begin{tabular}{@{}l r r r r r@{}}
\toprule
\thead{Bucket} & \thead{Queries} & \thead{Share} & \thead{BT median} & \thead{BT std} & \thead{Agent performance} \\
\midrule
Hard   & 73 & 33.2\% & $+12.96$ & 3.62 & 0.37 \\
Medium & 74 & 33.6\% & $+3.73$  & 1.85 & 0.63 \\
Easy   & 73 & 33.2\% & $-1.03$  & 3.23 & 0.80 \\
\bottomrule
\end{tabular}
\end{table}

We split the 220 \frontierfinance{} queries into \textbf{easy / medium / hard} terciles by BT score, with cutoffs at $\mathrm{BT} = +0.94$ and $+7.25$; Table~\ref{tab:difficulty-terciles} shows the three buckets are well separated. Note that difficulty is not simply a function of rubric count---the two correlate only moderately (Spearman $\rho = 0.71$). The examples in Table~\ref{tab:examples} illustrate that the key driver is the type of work required: easy queries resolve inside a single document even when the rubric is long (37 rubrics, all from one earnings call), whereas hard queries require cross-entity resolution, unbounded search, multi-hop reasoning, or causal analysis.

\begin{table}[t]
\centering\footnotesize\setlength{\tabcolsep}{4pt}
\caption{Example queries and rubric statistics for each difficulty bucket in \frontierfinance. Note that some queries shown have been significantly simplified to fit into the space.}
\label{tab:examples}
\begin{tabular}{@{}l r r p{5.9cm} p{4.6cm}@{}}
\toprule
\thead{Bucket} & \thead{BT} & \thead{Rubrics} & \multicolumn{1}{l}{\thead{Query (simplified)}} & \multicolumn{1}{l}{\thead{Main difficulty driver}} \\
\midrule
Easy   & $-9.9$  &   7 & ``What is the total employee count for US Steel Corporation?'' & Single fact, one entity, one filing \\
Easy   & $-5.7$  &  37 & ``What EPS guidance did Delta provide in its latest earnings call?'' & Many rubrics, but all from one earnings call \\
Medium & $+3.7$  &  21 & ``Which publicly traded US firms have exposure to robotics?'' & Unbounded entity scope, but shallow per entity \\
Medium & $+5.4$  &  19 & ``What have NEE, DUK, SO, and D said about load-growth expectations in the last year?'' & Multi-entity, multi-quarter transcript synthesis \\
Hard   & $+15.9$ &  21 & ``Identify Nike's newly announced FY2025 partner; then triangulate that partner's revenue exposure across Nike's, the partner's, and Adidas/UA filings.'' & Multi-hop conditional reasoning, cross-entity \\
Hard   & $+22.0$ & 300 & ``Screen \$10B+ biotechs with 2025 Phase-3 readouts, rank by analyst price-target dispersion, then decompose the leader's realized vs.\ implied volatility and 13F flows.'' & Unbounded screening + multi-source retrieval + computation \\
\bottomrule
\end{tabular}
\end{table}

\textbf{Correlation with agent performance and human effort.}\quad Table~\ref{tab:difficulty-terciles} also reports the performance of Samaya's agent system (defined in Section~\ref{sec:experiments}) averaged across each difficulty bucket. The BT score inversely correlates with system performance: qualification rate declines monotonically across easy, medium, and hard ($0.80 \to 0.63 \to 0.37$; Spearman $\rho = -0.47$). The BT score also correlates with human effort: using expert solve times from the 27 Finance Agent Benchmark v2 queries \citep{bigeard2025financeagentbenchmark}, the median time rises $20 \to 40 \to 45 \to 60$ minutes across BT score quartiles (Spearman $\rho = 0.67$), confirming the score tracks genuine task difficulty.

\begin{figure}[t]
\centering
\includegraphics[width=0.9\linewidth]{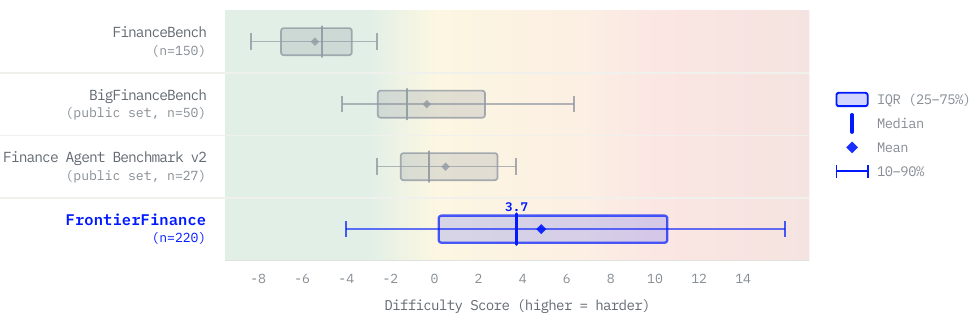}
\caption{Bradley--Terry difficulty distribution per benchmark on the shared consensus scale. The IQR box represents interquartile range. The background color bands mark \frontierfinance's easy/medium/hard buckets (cutoffs $+0.94$ / $+7.25$). \frontierfinance{} has the highest median and mean and by far the widest spread, reaching well into the hard band, while the three compared benchmarks sit mostly in the easy/medium range.}\label{fig:difficulty-comparison}
\end{figure}

\textbf{Comparison to existing finance benchmarks.}
We compare \frontierfinance{} against the three most similar public finance benchmarks---FinanceBench \citep{islam2023financebench}, BigFinanceBench \citep{wang2026bigfinancebench}, and Finance Agent Benchmark v2 \citep{bigeard2025financeagentbenchmark}---positioning all queries onto the same BT scale via anchor pairings (Appendix~\ref{sec:appendix_difficulty}). Figure~\ref{fig:difficulty-comparison} shows \frontierfinance{} is substantially harder and spans a wider range, with the highest median and widest spread; the three compared benchmarks cluster in the easy and medium ranges. Only 0--2\% of Finance Agent v2 and BigFinanceBench queries fall in the hard category, and FinanceBench---designed as single-fact 10-K lookups---is entirely easy. This gap reflects task design: \frontierfinance{} targets multi-source synthesis tasks rather than single-document lookups. Full statistics are in Appendix~\ref{sec:appendix_difficulty} (Tables~\ref{tab:boxplot} and~\ref{tab:positioning}).

\section{Experiment Setup}
\label{sec:experiments}

Frontier agentic AI systems have two essential parts: the LLM backbone, which decides what tool calls to make and what responses to generate, and the agent harness---the surrounding software (prompts, tools, execution environments, and the orchestration loop) through which the LLM perceives and acts on its environment.

With \frontierfinance, we benchmark frontier models and systems using three distinct types of harnesses:
\begin{itemize}
\item \textbf{Web search harness}: a minimal harness pairing each model with its built-in web search tool, providing a baseline for web-search-only performance.
\item \textbf{Finance Agent v2 harness}: an open-source harness connecting the model to six specialized tools: the SEC EDGAR API, a market price data API, web search, HTML parsing, long-HTML search, and a calculator \citep{bigeard2025financeagentbenchmark}. Included as a reproducible agent harness created specifically for finance tasks.
\item \textbf{In-house Samaya agent harness}: a production-grade harness combining Samaya's custom models, tools, data index, and retrieval engines, optimized for both quality and efficiency. Included to benchmark how a highly optimized finance harness performs.
\end{itemize}

\textbf{Implementation Details.}
For the web search harness, we evaluated Claude Opus 4.8, GPT 5.5, and Gemini 3.1 Pro paired with their built-in web search APIs, with reasoning effort at each model's API default.

For the Finance Agent v2 harness, we adapted its original implementation\footnote{\url{https://github.com/vals-ai/finance-agent-v2}} and made two changes. First, we re-implemented the agent orchestration with the LangChain library for maximum compatibility with existing models and frameworks. Second, we introduced tool call limits of 200 tool calls and 300 seconds, ensuring evaluation under a realistic finite budget. These limits are included in the system prompt; when either is hit, the model is prompted to provide a final response. We applied the same limits to the in-house Samaya agent harness for fair comparison. We include the system and user prompts in Appendix~\ref{appendix:finance-agent-prompt}.

For all model calls, we set the temperature to 1.0 wherever required. We enable thinking or reasoning mode and leave the reasoning effort at each model's API default, as listed in Table~\ref{tab:system-performance}. We leave all other decoding parameters, such as top $k$ or top $p$, to their API defaults. We include additional API endpoint information in Appendix~\ref{appendix:api-endpoint-details}.

\textbf{Query Date Handling.}
For all tested systems, we provide both the query and its associated query date as input, and prompt the system to anchor its research on that date.
For Samaya's in-house harness, we additionally apply the query date as a retrieval cut-off for any data indexed in-house.

\subsection{Evaluation Metrics}

We evaluate systems along three dimensions: answer quality, cost, and latency.

\textbf{Answer Quality}. We measure answer quality using the \textbf{Rubric Qualification Rate}. Let $N$ represent the total number of queries in \frontierfinance. Each query $q_i$ has a set of expert-authored binary rubrics $R_i = \{r_{i,1}, \dots, r_{i,M_i}\}$, with $M_i = |R_i|$. Given a system's answer $a_i$ to query $q_i$, a group of LLM judges evaluates each rubric independently and returns a binary verdict
\begin{equation}
    s(r_{i,j}, a_i) \in \{0, 1\},
\end{equation}
where $1$ indicates that $a_i$ satisfies $r_{i,j}$ and $0$ otherwise.

We define per-query qualification rate as the fraction of that query's rubrics satisfied by the answer:
\begin{equation}
    Q_i = \frac{1}{M_i} \sum_{j=1}^{M_i} s(r_{i,j}, a_i) \in [0, 1].
\end{equation}

At dataset level, we then calculate \textbf{Macro-averaged Rubric Qualification Rate} as the mean of the per-query rates over all queries and use it to represent a system's performance:
\begin{equation}
    \mathrm{R} = \frac{1}{N} \sum_{i=1}^{N} Q_i
    = \frac{1}{N} \sum_{i=1}^{N} \frac{1}{M_i} \sum_{j=1}^{M_i} s(r_{i,j}, a_i).
\end{equation}

We report two variants: $R_{\text{all}}$ over all rubrics and $R_{\text{must-have}}$ over the must-have subset. We use \emph{macro} over \emph{micro} averaging so that the metric rewards consistent performance across all queries, rather than excelling on a few high-rubric-count queries at the expense of others.

We score each rubric by majority vote across three independent LLM judges: GPT 5.4, Gemini 3.1 Pro, and Claude Sonnet 4.6. We chose this ensemble to avoid single-provider bias and because their majority votes closely matched a larger committee of nine judges in preliminary experiments. The judge prompt includes finance-specific instructions. We include the LLM judge prompt in Appendix~\ref{appendix:llm-judge-prompts} and open-source our grading pipeline for reproducibility.

\textbf{Latency}. We measure per-query latency as wall-clock time from query receipt to full answer, averaged over all queries.

\textbf{Cost}. We measure per-query cost as the API cost of the agentic LLM, excluding external APIs, storage, and data-index pricing for which precise estimates are hard to obtain. We account for cached and non-cached input tokens at their respective rates, crediting models that offer lower cached-token pricing.

\begin{table}[!t]
  \centering
  \small
  \caption{System performance on \frontierfinance{}, grouped by harness and ranked by rubric qualification rate within each group. $R_{\text{all}}$ represents macro-averaged qualification rate over all rubrics and $R_{\text{must-have}}$ represents qualification rate over the must-have subset of rubrics. \textit{Reasoning effort} is the API default for the underlying model. Overall best results are highlighted with boldface, and best results within each harness category are underlined. Average cost for Gemini 3.1 Pro under Web Search Harness is not shown because we were not able to obtain a fair estimate of the LLM cost from their search-grounded API.}
  \label{tab:system-performance}
  \begin{tabular}{@{}lccccc@{}}
    \toprule
    \textbf{System} & \textbf{Reasoning effort} & \textbf{$R_{\text{all}}$ (\%)} & \textbf{$R_{\text{must-have}}$ (\%)} & \textbf{Avg.\ latency (s)} & \textbf{Avg.\ cost (\$)} \\
    \midrule
    \multicolumn{6}{@{}l}{\textit{Web Search Harness}} \\
    \quad Claude Opus 4.8 & high & \underline{33.0} & \underline{40.6} & \phantom{0}53.9 & 1.47 \\
    \quad Gemini 3.1 Pro & high & 30.7 & 39.4 & \phantom{0}91.1 & -- \\
    \quad GPT 5.5 & medium & 20.7 & 26.5 & 192.1 & 0.55 \\
    \addlinespace
    \midrule
    \multicolumn{6}{@{}l}{\textit{Finance Agent v2 Harness}} \\
    \quad Claude Fable 5 & high & \underline{49.2} & \underline{57.6} & 164.5 & 4.06 \\
    \quad GPT 5.6 Sol & medium & 46.8 & 57.2 & 170.7 & 3.03 \\
    \quad Kimi K3 & max & 46.4 & 56.4 & 336.0 & 0.90 \\
    \quad Gemini 3.6 Flash & medium & 46.3 & 54.2 & 163.6 & 2.41 \\
    \quad Claude Opus 4.8 & high & 45.0 & 53.7 & 155.8 & 2.61 \\
    \quad GPT 5.5 & medium & 43.5 & 53.2 & 233.0 & 2.80 \\
    \quad GLM 5.2 & max & 42.8 & 50.1 & 296.5 & 0.63 \\
    \quad DeepSeek V4 Pro & high & 40.5 & 50.1 & 202.6 & 0.68 \\
    \quad Gemini 3.1 Pro & high & 30.5 & 39.7 & 221.9 & 1.72 \\
    \addlinespace
    \midrule
    \multicolumn{6}{@{}l}{\textit{Samaya In-house Harness}} \\
    \quad Samaya (high effort) & -- & \textbf{56.0} & \textbf{61.7} & 277.8  & 1.81 \\
    \quad Samaya & -- & 52.9 & 58.5 & 218.1 & 0.93 \\
    \bottomrule
  \end{tabular}
\end{table}

\section{Results}
\label{sec:results}

\begin{figure}[t]
  \centering
  \begin{subfigure}[t]{0.48\linewidth}
    \centering
    \includegraphics[width=\linewidth]{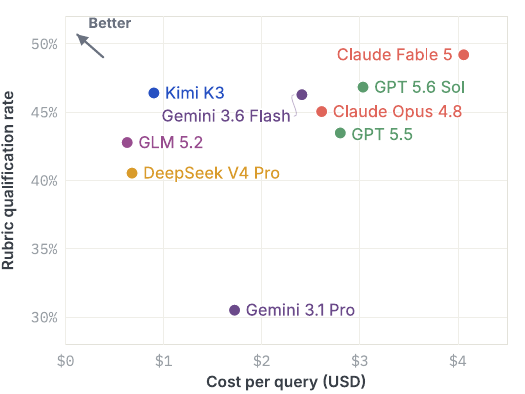}
    \label{fig:model-tradeoff-cost}
  \end{subfigure}
  \hfill
  \begin{subfigure}[t]{0.48\linewidth}
    \centering
    \includegraphics[width=\linewidth]{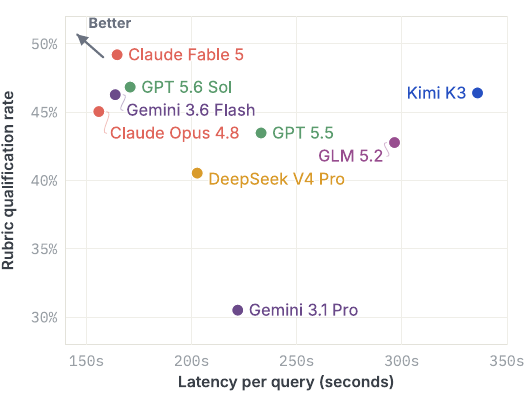}
    \label{fig:model-tradeoff-latency}
  \end{subfigure}
  \vspace{-1em}
  \caption{Model performance under the open-source Finance Agent v2 harness. Left: rubric qualification rate versus cost per query. Right: rubric qualification rate versus latency per query. In both figures, models toward the top-left offer a better quality-cost (or quality-latency) tradeoff.}
  \label{fig:model-tradeoffs}
\end{figure}

\begin{figure}[t]
\centering
\includegraphics[width=0.9\linewidth]{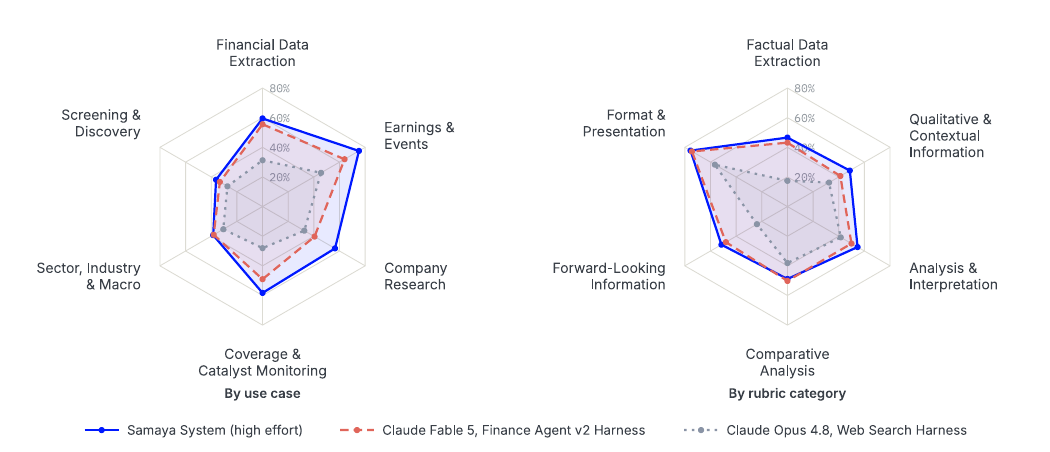}
\caption{Radar charts showing how system performance varies by use case (left) and rubric categories (right). For use case we report macro-averaged rubric qualification rate, as main results. For rubric breakdown we report micro-averaged rubric qualification rate. For readability, we only include best systems under each harness type.}
\label{fig:performance-breakdown}
\end{figure}

\begin{figure}[t]
\centering
\includegraphics[width=0.9\linewidth]{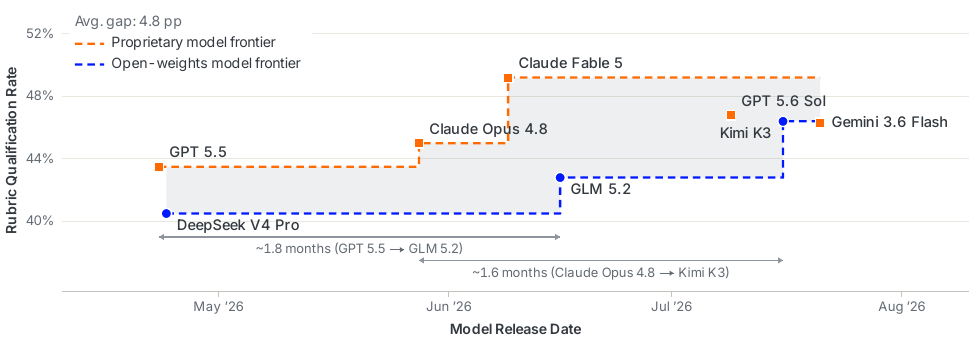}
\caption{Frontier performance on \frontierfinance{} from proprietary versus open-weight models against model release date. All models shown are evaluated under the Finance Agent v2 harness.}
\label{fig:model-timeline}
\end{figure}

\textbf{The harness shapes performance and cost more than the underlying model.}\quad Harness type is the dominant factor in system performance (Table~\ref{tab:system-performance}). The best system under the Samaya harness outperforms the best under the Finance Agent v2 harness, which in turn outperforms the best web search system---and this ordering holds across all six use cases and rubric categories (Figure~\ref{fig:performance-breakdown}). The Samaya harness also achieves its advantage at lower cost: Samaya (high effort) leads at 56\% while the best FA-v2 system (Claude Fable 5) reaches 49.2\%.

\textbf{Top proprietary models show non-linear quality--cost scaling but no latency penalty.}\quad Among proprietary models under the Finance Agent v2 harness, Claude Fable 5 leads at 49.2\%, followed closely by GPT 5.6 Sol at 46.8\% (Figure~\ref{fig:model-tradeoffs}). Quality and cost do not scale linearly: Fable 5 achieves a relative 9\% gain over Claude Opus 4.8 while incurring 56\% more cost, and a further 5\% gain over GPT 5.6 Sol at 34\% more cost. Notably, this quality advantage does not come with a latency penalty---both Fable 5 and GPT 5.6 Sol are among the fastest systems under this harness, a pattern we trace to their more efficient tool use in Section~\ref{sec:trajectory-analysis}.

\textbf{Open-weight models match proprietary quality at a fraction of the cost, within two months of release.}\quad Kimi K3, the best open-weight model, reaches 46.4\%---just 0.4pp behind GPT 5.6 Sol (46.8\%) and 2.8pp behind Claude Fable 5 (49.2\%)---at \$0.90 per query versus \$3.03 and \$4.06 respectively. Two of the four models on the quality--cost Pareto frontier are open-weight: GLM 5.2 (\$0.63, 42.8\%) and Kimi K3 (\$0.90, 46.4\%) offer the best quality-per-dollar among all systems, with no proprietary model matching their efficiency at comparable quality levels. Figure~\ref{fig:model-timeline} shows the gap is closing fast: GLM 5.2 nearly matched GPT 5.5 (42.8\% vs 43.5\%) within 1.8 months of its release, and Kimi K3 surpassed Claude Opus 4.8 (46.4\% vs 45.0\%) within 1.6 months, with an average open-to-proprietary gap of just 4.8 percentage points.

\begin{figure}[t]
  \centering
  \begin{subfigure}[t]{0.48\linewidth}
    \centering
    \includegraphics[width=\linewidth]{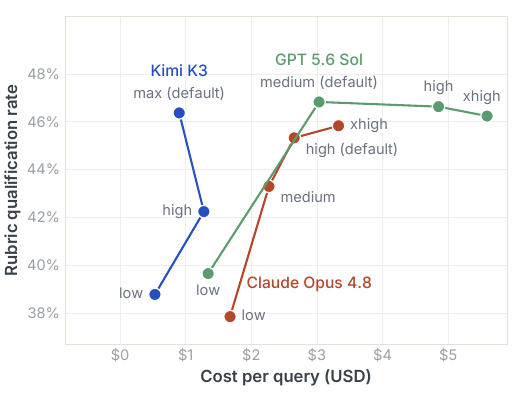}
    \label{fig:reasoning-effort-analysis}
  \end{subfigure}
  \hfill
  \begin{subfigure}[t]{0.48\linewidth}
    \centering
    \includegraphics[width=\linewidth]{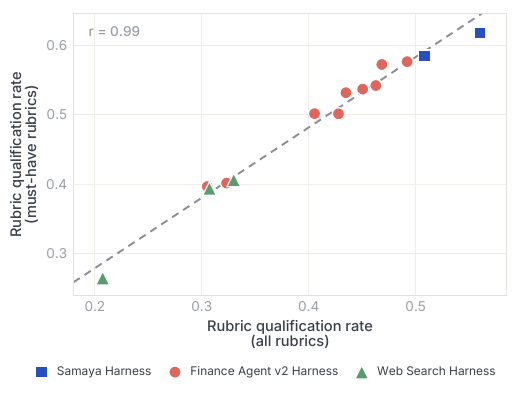}
    \label{fig:qualification-rate-correlation}
  \end{subfigure}
  \vspace{-1em}
  \caption{Further analysis of model performance under Finance Agent v2 harness. Left: performance of GPT 5.6 Sol, Opus 4.8 and Kimi K3 under varying reasoning effort settings. Right: correlation of model qualification rate for all rubrics and must-have rubrics under different system harnesses.}
  \label{fig:results-analysis}
\end{figure}

\textbf{Model performance scales with reasoning effort, but with diminishing returns.}\quad This holds across harnesses and models. Within the Samaya harness, the high-effort variant improves qualification rate by a relative 6\% over the default but at roughly 2$\times$ the cost. Among FA-v2 models, Figure~\ref{fig:results-analysis} (left) shows the same pattern: all three selected models produce higher qualification rates as effort increases, but returns diminish beyond each model's default level. GPT 5.6 Sol plateaus at \emph{medium} effort with a slight downward trend at higher settings. Kimi K3 shows an unconventional pattern: at \emph{max} effort, its qualification rate rises while its cost falls---higher reasoning induces more efficient tool use, reducing tool calls from 26.7 at \emph{high} to 19.6 at \emph{max}.

\textbf{Must-have and all-rubric qualification rates are near-perfectly correlated.}\quad All rubrics carry a boolean must-have label indicating whether an answer is significantly incomplete without that criterion. Figure~\ref{fig:results-analysis} (right) plots $R_{\text{all}}$ against $R_{\text{must-have}}$ for all systems. The two correlate at $r = 0.99$ across all evaluated systems, suggesting that $R_{\text{must-have}}$ is a reliable proxy for overall qualification rate.

\section{Model Trajectory Analysis}
\label{sec:trajectory-analysis}

We analyze model trajectories in depth to understand how behavioral differences drive efficiency and performance differences.
For consistency, we restrict this analysis to systems using the Finance Agent v2 harness.
Our analysis reveals the key findings below.

\textbf{Models vary sharply in parallelism and tool-call volume, yet top performers converge on similar latency.}\quad 
Table~\ref{tab:volume} presents overall trajectory statistics for various models.
Models differ most sharply in tools per turn: GPT 5.6 Sol batches most aggressively at 5.86 tools/turn, finishing in just 8.9 turns; Gemini 3.6 Flash is at the opposite extreme, issuing every call sequentially (1.00 tools/turn) across 25.9 turns---roughly $3\times$ longer than its peers. The two best-performing models sit at opposite ends of total tool-call volume: Claude Fable 5 (49.2\%) issues only 16.6 calls per query while GPT 5.6 Sol (46.8\%) issues 46.3---nearly $3\times$ more---suggesting that tool-call volume alone does not determine quality. Total output tokens vary considerably (3k--25k), but answer tokens are stable across top models at 1.6k--1.9k per query; the remainder is thinking and tool invocation tokens, with thinking tokens likely dominating since invocations are short by nature. Despite these differences in parallelism and output volume, Fable 5 (164.5s), GPT 5.6 Sol (170.7s), and Gemini 3.6 Flash (163.6s) post nearly identical latencies---suggesting the two strategies trade off in wall-clock time. We find no clear correlation between answer token length and qualification rate.

\begin{table}[t]
  \centering
  \small
  \caption{Trajectory statistics for different models under the Finance Agent v2 harness, ordered by qualification rate. $R_{\text{all}}$ is the macro-averaged qualification rate over all rubrics. \emph{Total output}, \emph{Answer tokens}, \emph{Tool calls}, and \emph{Turns} represent per-query mean statistics over all answered queries, where total output tokens include output tokens for thinking, tool calls and final answers, and each turn could have multiple parallel tool calls. \textit{Avg. tool / turn} is calculated by dividing total tool calls over total turns and then taking macro average, excluding the closing submit turn.}
  \label{tab:volume}
  \begin{tabular}{@{}lcccccc@{}}
    \toprule
    \textbf{System} & \textbf{$R_{\text{all}}$ (\%)} & \textbf{Total output} & \textbf{Answer tokens} & \textbf{Tool calls} & \textbf{Turns} & \textbf{Avg. tools / turn} \\
    \midrule
    Claude Fable 5   & 49.2 & 14{,}119 & 1{,}753 & 16.6 & \phantom{0}7.5 & 2.55 \\
    GPT 5.6 Sol      & 46.8 & 12{,}545 & 1{,}930 & 46.3 & \phantom{0}8.9 & 5.86 \\
    Kimi K3          & 46.4 & 15{,}828 & 1{,}854 & 19.6 & 10.4 & 2.09 \\
    Gemini 3.6 Flash & 46.3 & 24{,}820 & 1{,}946 & 24.8 & 25.9 & 1.00 \\
    Claude Opus 4.8  & 45.0 & 12{,}445 & 1{,}644 & 16.1 & \phantom{0}8.0 & 2.32 \\
    GPT 5.5          & 43.5 & 14{,}406 & 1{,}814 & 33.7 & 18.6 & 1.92 \\
    GLM 5.2          & 42.8 & 22{,}673 & 2{,}580 & 26.9 & 13.3 & 2.19 \\
    DeepSeek V4 Pro  & 40.5 & 11{,}464 & 2{,}162 & 41.1 & 21.3 & 2.03 \\
    Gemini 3.1 Pro   & 30.5 & \phantom{0}3{,}276 & \phantom{0,}859 & 16.1 & 15.6 & 1.10 \\
    \bottomrule
  \end{tabular}
\end{table}

\begin{figure}[!t]
\vspace{1em}
  \centering
  \begin{subfigure}[t]{0.4\linewidth}
    \centering
    \includegraphics[width=\linewidth]{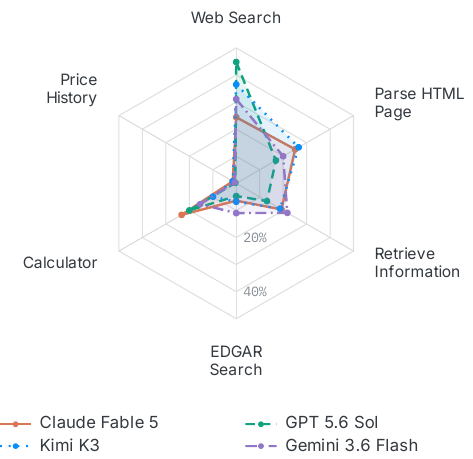}
    \label{fig:toolmix}
  \end{subfigure}
  \hspace{0.1\linewidth}
  \begin{subfigure}[t]{0.4\linewidth}
    \centering
    \includegraphics[width=\linewidth]{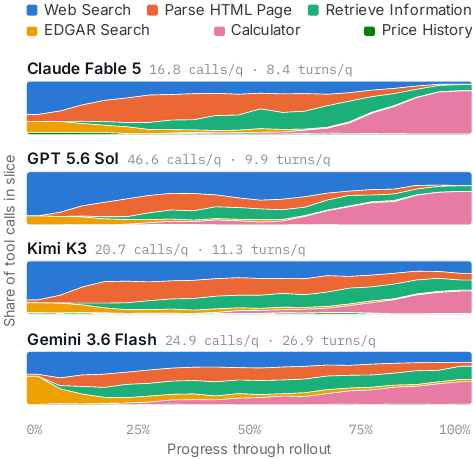}
    \label{fig:rollout}
  \end{subfigure}
  \vspace{-1em}
  \caption{Tool-use behavior under the Finance Agent v2 harness, for four top-performing models. Left: share of each system's tool calls going to each of the six tools. Right: evolution of the tool mix over the rollout, normalized to each query's own rollout length.}
  \label{fig:tool-behavior}
\end{figure}

\begin{figure}[t]
  \centering
  \begin{subfigure}[t]{0.4\linewidth}
    \centering
    \includegraphics[width=\linewidth]{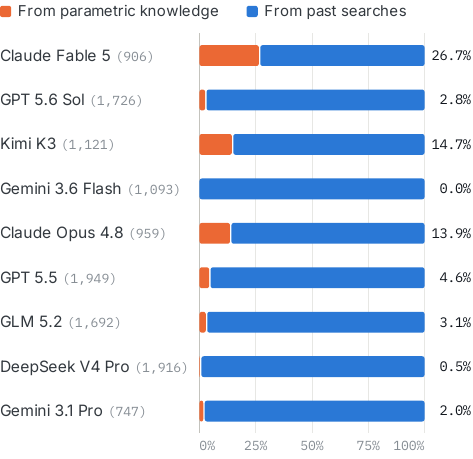}
    \label{fig:provenance-split}
  \end{subfigure}
  \hspace{0.1\linewidth}
  \begin{subfigure}[t]{0.4\linewidth}
    \centering
    \includegraphics[width=\linewidth]{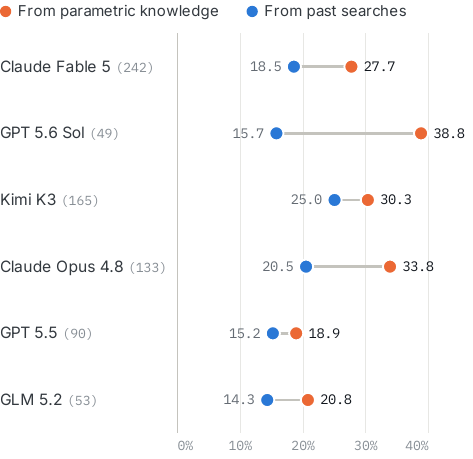}
    \label{fig:provenance-errrates}
  \end{subfigure}
  \vspace{-1em}
  \caption{Origin of the links each system attempts to crawl and parse with the \texttt{parse\_html\_page} tool and the distribution of URL parse errors resulting from it. Left: share of URL links coming from parametric knowledge rather than past search tool calls (total URL attempts in brackets). Right: parse error rate by link origin, for systems with more than 2.5\% of parses from parametric knowledge (parametric-knowledge parse count in brackets).}
  \label{fig:provenance}
\end{figure}

\textbf{Systems share a similar tool mix, with characteristic model-family differences.}\quad Web search dominates tool usage in nearly every system, followed by page parsing and retrieval; price history is negligible throughout (Figure~\ref{fig:tool-behavior} left; full split in Appendix~\ref{appendix:toolsplit}). Within this common shape, model-family patterns emerge: GPT 5.6 Sol directs 44.8\% of calls to web search, the highest of the four; Gemini systems and GLM 5.2 allocate the most to EDGAR full-text search; and Claude Fable 5's most-used tool is page parsing.

\textbf{Tool use follows a common three-phase trajectory.}\quad Figure~\ref{fig:tool-behavior} (right) plots tool-call composition over each system's rollout, normalized to query length. All four systems open with a data-gathering phase in which web and EDGAR search account for at least 80\% of calls in the first 10\% of the rollout. A research phase follows, with page parsing and corpus retrieval dominating mid-rollout activity as systems read the sources their searches surfaced. The final phase is answer preparation, marked by a sharp shift toward the calculator---for Fable 5, 80\% of calls in the last 10\% of the rollout. This structure holds irrespective of rollout length and call volume, suggesting the phase boundaries reflect task structure more than any one model's policy. Systems differ mainly in how decisively they transition: Fable 5 shows the sharpest phase separation, while GPT 5.6 Sol continues issuing web searches well into the final quarter.

\textbf{Some top-performing models recall sources from parametric knowledge, sidestepping discovery.}\quad
We track the origin of every HTML page a model attempts to crawl and parse, classifying each link as coming from a \emph{past search} (its domain appeared previously in the same trace) or from the model's \emph{parametric knowledge} (the domain appears without prior search history).
The two Claude models and Kimi K3 draw on parametric knowledge at substantially higher rates than the rest: 26.7\% of Fable 5's parses target self-produced domains, roughly half that rate for Kimi K3 and Claude Opus 4.8, while every other system sits below 5\% (Gemini 3.6 Flash produces zero across 1{,}093 parses; Figure~\ref{fig:provenance} left). 
These models navigate directly to canonical financial sources---\texttt{sec.gov}, \texttt{fred.stlouisfed.org}, \texttt{macrotrends.net}---rather than discovering them through search.
While this may appear as a good property of the model, we uncover a particular failure pattern that results from it: URLs from parametric knowledge incur significantly higher access error rates---due to hallucinated URLs or pages blocking crawling---than search-discovered URLs (Figure~\ref{fig:provenance} right), causing token waste and context pollution.

\section{Limitations}
FrontierFinance anchors each query to a specific date (Section~\ref{sec:timestamped-annotation}), which ensures reproducible evaluation today but introduces a temporal limitation as models advance. As future LLMs are trained on data postdating the query dates, they may answer from parametric knowledge rather than through active retrieval and tool use---the very capabilities the benchmark is designed to test. This risk mirrors the failure pattern we observe in current models (Section~\ref{sec:trajectory-analysis}), but will grow more acute over time. Periodic re-annotation with newer query dates is one way to mitigate this issue.

Use cases such as \emph{Screening \& Discovery} are inherently subjective: two analysts might produce different yet equally valid answers.
A system whose answer overlaps with the rubric may outscore one with an equally correct but differently framed response.
We note, however, that this limitation is mitigated by statistical power: averaged across a large number of uncorrelated query-rubric pairs, a higher qualification rate still indicates stronger alignment with the analysis process that reflects financial best practices.
The evaluation is therefore fair at scale even where individual queries are subjective. In this work we spread the annotation budget over more queries; future work could collect multiple rubrics per subjective query to score systems more precisely.

\section{Conclusion}
We presented \frontierfinance, an open benchmark for evaluating AI agents on professional investment research. By spanning six use cases across the full investor workflow and scoring long-form answers against expert-authored, source-attributed rubrics, \frontierfinance{} measures capabilities that existing finance benchmarks largely overlook. Our evaluation shows that the benchmark is broad and difficult, that the tool harness shapes performance as much as the underlying model, and that substantial headroom remains, particularly on use cases requiring broad screening and macro-oriented research. We release the dataset and grading code, and we hope \frontierfinance{} serves as a shared standard for measuring progress on finance AI agents. As models and systems advance, we plan to expand the benchmark from our larger internal annotation pool and to report results on new systems over time.

\section{Acknowledgement}

We thank Christos Baziotis, Jack Hessel, Jack Santos Silva and Mingyi Yang for their contribution to early data collection process, and Suharsh Sivakumar, Kyle Chang and Bram Mulders for providing engineering and infrastructure support.

\bibliographystyle{unsrtnat}
\bibliography{main}

\newpage
\appendix

\section{Sample queries and rubrics by use case}\label{appendix:samples}

The following six examples are the queries featured on the \frontierfinance{} benchmark page, one per use case.
For each query we show the first seven rubric items as displayed on the benchmark page (must-have rubrics marked \textbf{M}; supplementary marked \textbf{S}), along with the total rubric count.
The complete set of 220 queries and all 11,543 rubrics is available via the Hugging Face data viewer at \url{https://huggingface.co/datasets/samaya-ai/FrontierFinance}.

\subsection*{Screening \& Discovery}

\textbf{Query:} Which MedTech companies stand to gain from the new US presidential regime in 2025, especially regarding tariffs and manufacturing flexibility, and which are expected to be negatively impacted? \hfill\textit{Query date: 2025-05-24}

\medskip
\begin{itemize}
  \item[\textbf{S}] Summarizes all the information relating to the impact of the new US presidential regime in 2025 on MedTech companies as bullet points under the headers of respective company names.
  \item[\textbf{M}] Provides information about Becton Dickinson and Co (BDX) stating that the company is expected to not benefit under the Trump Administration of 2025.
  \item[\textbf{S}] Provides information about Becton Dickinson and Co (BDX) stating that the company is known for developing and manufacturing a wide range of medical devices, instrument systems, and reagents used in various healthcare settings.
  \item[\textbf{S}] Provides information about Becton Dickinson and Co (BDX) stating that the company reported total sales of USD 5.27 billion for Q1 2025.
  \item[\textbf{S}] Provides information about Becton Dickinson and Co (BDX) stating that the company cut its 2025 profit forecast.
  \item[\textbf{S}] Provides information about Becton Dickinson and Co (BDX) stating that the company expects its 2025 profit per share to be between USD 14.06 and USD 14.34.
  \item[\textbf{M}] Provides information about Becton Dickinson and Co (BDX) stating that the company cut its 2025 profit forecast due to potential tariff hit.
  \item[$\cdots$] \textit{+27 more rubric items}
\end{itemize}

\subsection*{Company Research}

\textbf{Query:} What has driven the slowdown in Intel's revenues over the past two years? \hfill\textit{Query date: 2025-05-18}

\medskip
\begin{itemize}
  \item[\textbf{M}] States that Intel Corporation (INTC) reported total revenue of USD 54.20 billion in 2023.
  \item[\textbf{M}] States that Intel Corporation (INTC) reported total revenue of USD 53.10 billion in 2024.
  \item[\textbf{S}] States that Intel Corporation (INTC) reported a 5.00\% decline in notebook volume in 2023.
  \item[\textbf{S}] States that Intel Corporation (INTC) reported a 9.00\% decline in desktop volume in 2023.
  \item[\textbf{M}] States that Intel Corporation (INTC) experienced a decline in client notebook volume due to weak consumer demand in 2023.
  \item[\textbf{M}] States that Intel Corporation (INTC) experienced a decline in client desktop volume due to soft education and small and medium-sized business (SMB) markets in 2023.
  \item[\textbf{M}] States that Intel Corporation (INTC) reported a reduction in average selling prices for notebooks in 2023.
  \item[$\cdots$] \textit{+33 more rubric items}
\end{itemize}

\subsection*{Sector, Industry \& Macro}

\textbf{Query:} How do room nights booked and revenue compare across Booking Holdings, Expedia, Airbnb, and other top Online Travel Agencies (OTAs)? \hfill\textit{Query date: 2025-03-29}

\medskip
\begin{itemize}
  \item[\textbf{M}] States that the other top Online Travel Agencies are Trip.com Group Limited (TCOM) and MakeMyTrip Limited (MMYT) based on their market capitalizations of more than USD 10 billion.
  \item[\textbf{S}] Summarizes all the information relating to revenue and the number of nights booked for Booking Holdings Inc.\ (BKNG), Expedia Group, Inc.\ (EXPE), Airbnb, Inc.\ (ABNB), Trip.com Group Limited (TCOM) and MakeMyTrip Limited (MMYT) in a table format.
  \item[\textbf{M}] Provides information for Booking Holdings Inc.\ (BKNG) stating that the total room nights for full year 2024 were 1,144 million.
  \item[\textbf{M}] Provides information for Booking Holdings Inc.\ (BKNG) stating that the total room nights for full year 2023 were 1,049 million.
  \item[\textbf{M}] Provides information for Booking Holdings Inc.\ (BKNG) stating that the total room nights for full year 2022 were 896 million.
  \item[\textbf{S}] Provides information for Booking Holdings Inc.\ (BKNG) stating that the total room nights for full year 2024 increased by 9\% year over year.
  \item[\textbf{S}] Provides information for Booking Holdings Inc.\ (BKNG) stating that the total room nights for full year 2023 increased by 17\% year over year.
  \item[$\cdots$] \textit{+56 more rubric items}
\end{itemize}

\subsection*{Financial Data Extraction}

\textbf{Query:} Provide a breakdown of BP's upstream, downstream, and integrated gas and renewables segments' operating metrics over the past 6 years. \hfill\textit{Query date: 2025-10-09}

\medskip
\begin{itemize}
  \item[\textbf{S}] Presents BP p.l.c.\ (BP) operating metrics from 2019 to 2024 in a structured or tabular format.
  \item[\textbf{S}] Presents a footnote below the table for BP p.l.c.\ (BP) stating that for 2022, the Biogas supply volumes exclude Archaea Energy.
  \item[\textbf{M}] Provides information for BP p.l.c.\ (BP) stating that the Upstream Production for the year 2024 was 2.4~mmboe/d (million barrels of oil equivalent per day).
  \item[\textbf{M}] Provides information for BP p.l.c.\ (BP) stating that the Upstream Production for the year 2023 was 2.3~mmboe/d.
  \item[\textbf{M}] Provides information for BP p.l.c.\ (BP) stating that the Upstream Production for the year 2022 was 2.3~mmboe/d.
  \item[\textbf{M}] Provides information for BP p.l.c.\ (BP) stating that the Upstream Production for the year 2021 was 2.2~mmboe/d.
  \item[\textbf{M}] Provides information for BP p.l.c.\ (BP) stating that the Upstream Production for the year 2020 was 2.4~mmboe/d.
  \item[$\cdots$] \textit{+69 more rubric items}
\end{itemize}

\subsection*{Coverage \& Catalyst Monitoring}

\textbf{Query:} In what ways have the AI strategies of Alphabet and Microsoft diverged across the past 8 quarters? \hfill\textit{Query date: 2025-06-09}

\medskip
\begin{itemize}
  \item[\textbf{M}] States that Alphabet Inc (GOOGL) defines Artificial Intelligence (AI) as a profound platform shift central to its mission in 2024.
  \item[\textbf{M}] States that Alphabet Inc (GOOGL) launched Gemini 1 and Gemini 1.5 in 2024 as next-generation AI models.
  \item[\textbf{S}] States that Alphabet Inc (GOOGL) designed Gemini models to process and combine text, images, audio, video, and code in 2024.
  \item[\textbf{M}] States that Alphabet Inc (GOOGL) integrated Gemini models into Search, Ads, Chrome, Gmail, Maps, and YouTube in 2024.
  \item[\textbf{S}] States that Alphabet Inc (GOOGL) enabled Gemini models to serve billions of users across its core product suite in 2024.
  \item[\textbf{M}] States that Alphabet Inc (GOOGL) uses Vertex AI in Google Cloud to help developers build and scale generative AI applications in 2024.
  \item[\textbf{S}] States that Alphabet Inc (GOOGL) incorporates Gemini and Duet AI into Google Workspace to enhance productivity tools in 2024.
  \item[$\cdots$] \textit{+46 more rubric items}
\end{itemize}

\subsection*{Earnings \& Events}

\textbf{Query:} What positive and negative aspects came out of ABBV's last earnings call? \hfill\textit{Query date: 2025-05-01}

\medskip
\begin{itemize}
  \item[\textbf{S}] Clearly separates the positive points and negative points from AbbVie Inc.'s (ABBV) last earnings call transcript.
  \item[\textbf{S}] States that AbbVie Inc.\ (ABBV)'s latest earnings call was for Q1 2025.
  \item[\textbf{M}] Provides a ``Negative Point'' for AbbVie Inc.\ (ABBV) from its Q1 2025 earnings call transcript stating that global sales of Humira were down 49.5\% on an operational basis.
  \item[\textbf{S}] Provides a ``Negative Point'' for AbbVie Inc.\ (ABBV) stating that global sales of Humira were down due to faster share erosion from biosimilar competition.
  \item[\textbf{M}] Provides a ``Negative Point'' for AbbVie Inc.\ (ABBV) stating that aesthetics global sales were down 10.2\% on an operational basis.
  \item[\textbf{S}] Provides a ``Negative Point'' for AbbVie Inc.\ (ABBV) stating that Botox cosmetic revenues were down 10.7\%.
  \item[\textbf{S}] Provides a ``Negative Point'' for AbbVie Inc.\ (ABBV) stating that Juvederm sales were down 20\%.
  \item[$\cdots$] \textit{+29 more rubric items}
\end{itemize}

\section{Dataset statistics}\label{sec:appendix_datastats}

\textbf{Use cases.}\quad Each query carries exactly one of six use cases spanning the
investor workflow; Table~\ref{tab:usecases} gives the full counts and descriptions.

\begin{table}[h]
\centering\footnotesize\setlength{\tabcolsep}{4pt}
\caption{Use case descriptions and distribution.}\label{tab:usecases}
\begin{tabular}{@{}l r r p{6.2cm}@{}}
\toprule
\thead{Use case} & \thead{Queries} & \thead{Share} & \multicolumn{1}{l}{\thead{Description}} \\
\midrule
Financial Data Extraction   & 70 & 31.8\% & Quantitative data extraction and modeling from regulatory filings \\
Sector, Industry \& Macro   & 38 & 17.3\% & Sector, industry, and macroeconomic research across many entities \\
Earnings \& Events  & 36 & 16.4\% & Queries centered on a corporate communication event (earnings call or release, investor day, 8-K) \\
Company Research   & 32 & 14.5\% & Company-level research: business, strategy, and operations \\
Coverage \& Catalyst Monitoring  & 27 & 12.3\% & Tracking a name or topic over time across sources \\
Screening \& Discovery  & 17 &  7.7\% & Open-ended screening and discovery over an unbounded set of entities \\
\bottomrule
\end{tabular}
\end{table}

\textbf{Capabilities.}\quad Each query is additionally tagged with the reasoning and
retrieval capabilities it requires. The raw counts and
definitions for these capability categories are provided below in Table~\ref{tab:capabilities}.

\begin{table}[h]
\centering\footnotesize\setlength{\tabcolsep}{4pt}
\caption{Capability descriptions and distribution (a query may carry several, so counts sum to more than 220).}\label{tab:capabilities}
\begin{tabular}{@{}l r r p{6.2cm}@{}}
\toprule
\thead{Capability} & \thead{Queries} & \thead{Share} & \multicolumn{1}{l}{\thead{Description}} \\
\midrule
Qualitative Synthesis              & 99 & 45.0\% & Curate and synthesize qualitative material across documents \\
Exhaustive Retrieval\,/\,Temporal  & 71 & 32.3\% & Retrieve every qualifying item across multiple time periods \\
Exhaustive Retrieval\,/\,Cross-Entity & 44 & 20.0\% & Retrieve every qualifying item across multiple entities \\
Exhaustive Retrieval\,/\,Thematic  & 32 & 14.5\% & Collect every instance of a theme or category \\
Numerical Reasoning                & 59 & 26.8\% & Compute or derive figures from multiple inputs \\
Multi-hop Workflows               & 53 & 24.1\% & Chained, conditional, multi-step tasks \\
Causal Reasoning                   & 14 &  6.4\% & Explain the drivers behind an outcome \\
Simple Retrieval                   & 38 & 17.3\% & Single-fact lookup or verification \\
\bottomrule
\end{tabular}
\end{table}

\textbf{Rubric content categories.}\quad Each of the 11{,}543 public-benchmark rubrics is
classified into one of eight content categories describing the kind of content the
rubric demands (Table~\ref{tab:rubriccats}). Factual data extraction is the plurality
(74\%), consistent with rubrics being written as objective, checkable criteria; the
remaining quarter spans the qualitative, forward-looking, analytical, and comparative
content that a complete answer must also deliver.

\begin{table}[h]
\centering\footnotesize\setlength{\tabcolsep}{4pt}
\caption{Rubric category descriptions and distribution.}\label{tab:rubriccats}
\begin{tabular}{@{}l r r p{6.6cm}@{}}
\toprule
\thead{Rubric category} & \thead{Rubrics} & \thead{Share} & \multicolumn{1}{l}{\thead{Description}} \\
\midrule
Factual Data Extraction              & 8,547 & 74.0\% & Specific, verifiable data points (values, dates, names) from primary sources \\
Qualitative \& Contextual Information & 1,039 &  9.0\% & Descriptive, non-numeric content: definitions, descriptions, commentary \\
Forward-Looking Information           &   814 &  7.1\% & Future-oriented content: guidance, forecasts, projections \\
Analysis \& Interpretation            &   558 &  4.8\% & Synthesis, causal reasoning, and interpretive insight \\
Comparative Analysis                  &   264 &  2.3\% & Side-by-side comparison of entities, metrics, or periods \\
Format \& Presentation                &   131 &  1.1\% & How information is structured or presented \\
Source \& Methodology                 &   112 &  1.0\% & Required sources, research scope, or calculation method \\
Inquiry \& Question Generation        &    78 &  0.7\% & Formulating questions for management or further research \\
\bottomrule
\end{tabular}
\end{table}

\textbf{Rubric data source categories.}\quad
In our annotation guideline, we define a ten-category data source taxonomy.
Each top-level data source category is paired with examples and a finer set of second-level categories for clarity (e.g., 10-K or 10-Q as second-level categories for company filings).
At annotation time, we ask the annotator to record the data source categories that each written rubric demands.
For example, if the data in a specific rubric is drawn from a company 10-K document, the rubric should be tagged as using company filing as top-level source and 10-K as second-level source.
We include this data source tag in the final released dataset, and show their detailed distributions by use cases in Table~\ref{tab:rubric-data-source}.

\begin{table}[t]
  \centering
  \caption{Distribution of rubric data sources by use case. Each cell gives the number of rubric items drawn from that data source, with that source's share of the row in parentheses (\%); \emph{Total} is the number of rubric items for that use case. Shares are rounded to whole numbers summing to 100 within each row. Column headings abbreviate the data sources: SEC = SEC filings; Company = company originated content (e.g., transcripts, investor presentations, press releases); Professional\ = professional knowledge; Market = market price data; News = news and media sources; Regulatory\ = regulatory and legal data; Other = other categories.}
  \label{tab:rubric-data-source}
  \footnotesize
    \setlength{\tabcolsep}{3pt}
    \renewcommand{\arraystretch}{1.1}
    \begin{tabular}{@{}lrrrrrrrr@{}}
    \toprule
    Use case & SEC & Company & Professional & Market & News & Regulatory & Other & Total \\
    \midrule
    Financial Data Extraction & 3,378 (59) & 1,222 (21) & 863 (15) & 97 (2) & 28 (0) & 2 (0) & 167 (3) & 5,757 \\
    Sector, Industry \& Macro & 643 (25) & 378 (15) & 825 (32) & 334 (13) & 78 (3) & 26 (1) & 269 (11) & 2,553 \\
    Screening \& Discovery & 81 (8) & 107 (11) & 194 (19) & 291 (29) & 120 (12) & 97 (10) & 106 (11) & 996 \\
    Earnings \& Events & 65 (8) & 568 (68) & 135 (16) & 12 (2) & 1 (0) & 9 (1) & 44 (5) & 834 \\
    Coverage \& Catalyst & 136 (19) & 408 (57) & 95 (13) & 54 (8) & 3 (0) & 0 (0) & 24 (3) & 720 \\
    Company Research & 235 (34) & 214 (31) & 108 (16) & 54 (8) & 38 (6) & 0 (0) & 34 (5) & 683 \\
    \midrule
    \textbf{All use cases} & 4,538 (39) & 2,897 (25) & 2,220 (19) & 842 (7) & 268 (3) & 134 (1) & 644 (6) & 11,543 \\
    \bottomrule
    \end{tabular}
\end{table}

\section{Difficulty analysis details}\label{sec:appendix_difficulty}

\subsection{Methodology}

\textbf{What the score measures.}\quad Our difficulty score is a Bradley--Terry (BT)
latent score \citep{bradley1952rank} fit over pairwise ``which query is harder?''
judgments across five reasoning axes (retrieval breadth, reasoning depth, entity
scope, time scope, qualitative ambiguity) on Samaya's internal pool ($n = 4{,}212$).
We estimate latent difficulties $\theta$ by minimizing the confidence-weighted,
regularized negative log-likelihood
\[
  \hat{\theta} \;=\; \argmin_{\theta}\; -\sum_{k} c(m_k)\,\log \sigma\!\big(\theta_{w_k}-\theta_{\ell_k}\big)\;+\;\lambda\lVert\theta\rVert_2^2,
\]
where $\theta_i$ is query $i$'s difficulty, $(w_k,\ell_k)$ are the winner/loser query
in judgment $k$ for difficulty, $m_k$ its confidence margin, $\sigma$ the logistic function,
$c(1,2,3)=(0.3,\,1.0,\,1.3)$ the margin weights, and $\lambda = 10^{-3}$.
To place external benchmarks on the same scale, we pair their 227 queries with
Samaya-dataset anchors into 6,810 comparison pairs, score them with the judges,
and include them in the BT fit; each external query's percentile and bucket
are then read off the shared scale.

\textbf{Judging protocol and consensus fit.}\quad For every query in a pair, the judge sees
the query text, its date, use-case and capability tags, and the full criteria rubric
(each criterion labelled by essentiality, content type, and expected data source). The
prompt defines difficulty as the retrieval and cognitive effort an analyst would need for a
complete, defensible answer, tells the judge to weigh the five axes above, and warns
against difficulty proxies---rubric count, query length, single-document extraction
breadth---so a long but shallow query is not scored as hard. The judge returns a winner
(ties allowed but discouraged), a 1--3 confidence margin, a per-axis vote, and a brief
rationale; pairs are judged independently. Three judges---Gemini 3.1 Flash Lite,
Gemini 3 Flash, and Claude Sonnet 4.5---score every pair; we merge them \emph{per pair} by
signed-average (the pair is a tie if the mean is within $0.5$), then fit the single
confidence-weighted Bradley--Terry model above to the resulting consensus verdicts---each
judgment weighted by its margin at $c(1,2,3)=(0.3,\,1.0,\,1.3)$ so close calls count for a
fraction of clear ones.

\subsection{Robustness}

\textbf{Robustness to the judge model.}\quad Fit separately, each judge's own ratings
reproduce the consensus approximately (Table~\ref{tab:judgemodel}): Spearman
$0.95$--$0.97$ on the scores and 83--88\% easy/medium/hard label agreement. 
We also observed that Claude Sonnet is the steadiest judge---it flips the fewest verdicts under order swap and ties most often.

\begin{table}[h]
\centering\small
\caption{Each judge's individual BT fit vs.\ the three-judge consensus, over the full pairwise set.}\label{tab:judgemodel}
\begin{tabular}{@{}l r r@{}}
\toprule
\thead{Judge} & \thead{BT-score Spearman $\rho$ vs.\ consensus} & \thead{Bucket-label agreement} \\
\midrule
Gemini 3 Flash        & 0.974 & 87.8\% \\
Gemini 3.1 Flash Lite & 0.955 & 83.1\% \\
Claude Sonnet 4.5     & 0.946 & 82.5\% \\
\bottomrule
\end{tabular}
\end{table}

\textbf{Position bias.}\quad Re-judging a fixed 20{,}000-pair sample with the two queries in
swapped A/B order flips the winner on 7\% of consensus results, concentrated in
low-confidence calls---28/11/2\% at margins 1/2/3 (Table~\ref{tab:flip}). Because the
confidence-weighted fit already discounts those close calls (margin-1 judgments enter at
weight 0.3), the result barely changes: Refitting on the swapped verdicts preserves the
ranking (Spearman $0.996$) and 95.4\% of easy/medium/hard labels.

\begin{table}[h]
\centering\small
\caption{A/B position-swap flip rate by the judge's own confidence margin, per judge and for the consensus (20{,}000-pair sample).}\label{tab:flip}
\begin{tabular}{@{}l r r r r@{}}
\toprule
\thead{Judge} & \thead{Margin 1 (close)} & \thead{Margin 2 (clear)} & \thead{Margin 3 (very clear)} & \thead{Overall} \\
\midrule
Gemini 3.1 Flash Lite & 44\% & 28\% & 5\% & 21\% \\
Gemini 3 Flash        & 33\% & 17\% & 3\% & 14\% \\
Claude Sonnet 4.5     & 27\% & 12\% & 2\% & 11\% \\
Consensus             & 28\% & 11\% & 2\% &  7\% \\
\bottomrule
\end{tabular}
\end{table}

\begin{figure}[h]
\centering
\includegraphics[width=0.6\linewidth]{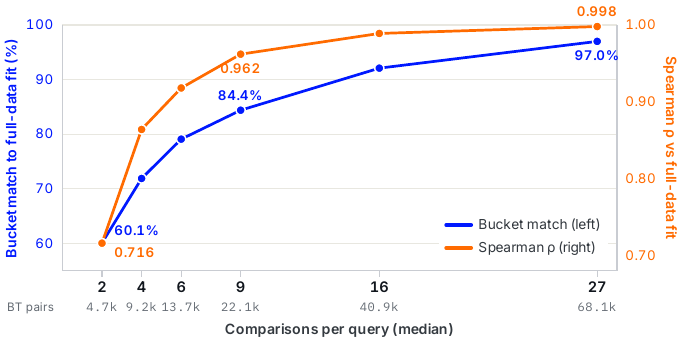}
\caption{Difficulty-scale stability vs.\ judgment volume. Bucket match (left axis) and Spearman $\rho$ (right axis) of a subsampled BT fit against the full-data consensus fit, as a function of comparisons per query (median; total BT pairs in parentheses). Both saturate well before the full budget---the ranking is essentially locked by $\sim$16 comparisons per query.}\label{fig:saturation}
\end{figure}

\textbf{Robustness to judgment volume.}\quad We probe the judgment budget two ways. First,
\emph{per query}: The full set averages about 35 comparisons per query, so we subsample to
$k$ comparisons each, refit, and compare to the full-data fit
(Fig~\ref{fig:saturation}); returns diminish quickly---the ranking is essentially locked
by $\sim$16 comparisons per query (Spearman $0.99$) and tercile labels reach 97\% agreement
by $\sim$27. Second, \emph{overall}: Refitting on random 50\% and 75\% subsamples of all
judgments (three draws each) reproduces the full-data scores at Spearman $0.99$ and
$0.996$ and flips only 8.0\% and 4.3\% of labels, respectively. 
These results show that we have sufficient pairwise samples to assign reliable scores.

\textbf{Per-query score uncertainty.}\quad We draw $B = 20$ bootstrap resamples of the
judgment set (sampling pairwise judgments with replacement to the original size), refit
the BT model on each, and take the standard deviation of a query's score across the refits
as its uncertainty. Scores are well-determined: On a scale spanning $\sim$36 units
($-14$ to $+22$), the median per-query standard deviation is 0.67 and the 90th percentile
1.37 (2--4\% of the range). About one query in five sits within one standard deviation of
an easy/medium or medium/hard cutoff, so near-boundary labels should be read as soft.

\subsection{Analysis with resulting difficulty scores}

\textbf{Difficulty is not criteria count.}\quad Rubric count only moderately predicts
difficulty (Spearman $\rho = 0.712$, $\rho^2 \approx 0.51$), explaining about half of the
rank variance; the depth and breadth of reasoning a query demands matter more than its raw
rubric count.

\textbf{Which axes drive the judgment.}\quad Restricting to non-tie pairs and pooling all
three judges, reasoning depth and retrieval breadth agree with the overall winner most
often (88.0\% and 85.0\%), followed by qualitative ambiguity (74.2\%), while time scope
(53.0\%) and entity scope (35.4\%) agree least---mainly because those two axes are
themselves tied far more often (entity 62\%, time 43\% of pairs, versus 12--15\% for the
two dominant axes). Retrieval and reasoning demands discriminate between queries far more
often than entity or temporal scope.

\textbf{Per-use-case difficulty.}\quad Difficulty varies systematically by use case
(Table~\ref{tab:usecasedifficulty}). Screening \& discovery and sector/industry \&
macro are the two hardest use cases across the board---open-ended, unbounded-scope
tasks that demand synthesis over many sources---while earnings \& events and company
research are comparatively more tractable.

\begin{table}[h]
\centering\small
\caption{Per-use-case difficulty, hardest to easiest.}\label{tab:usecasedifficulty}
\begin{tabular}{@{}l r r l@{}}
\toprule
\thead{Use case} & \thead{Queries} & \thead{Median difficulty pctile} & \thead{Hard / Med / Easy} \\
\midrule
Sector, Industry \& Macro        & 38 & 98.3 & 28 / 7 / 3 \\
Screening \& Discovery           & 17 & 93.9 & 10 / 4 / 3 \\
Financial Data Extraction      & 70 & 86.1 & 23 / 31 / 16 \\
Coverage \& Catalyst Monitoring & 27 & 68.2 & 5 / 9 / 13 \\
Earnings \& Events               & 36 & 52.7 & 4 / 11 / 21 \\
Company Research                   & 32 & 47.5 & 3 / 12 / 17 \\
\bottomrule
\end{tabular}
\end{table}

\textbf{External benchmarks.}\quad Tables~\ref{tab:boxplot} gives the full box-plot statistics for the three external benchmarks 
discussed in Section~\ref{sec:comparison}.

\begin{table}[h]
\centering\footnotesize\setlength{\tabcolsep}{5pt}
\caption{Bradley--Terry difficulty distribution per benchmark (box-plot statistics; higher = harder).}\label{tab:boxplot}
\begin{tabular}{@{}l r r r r r r r r@{}}
\toprule
\thead{Benchmark} & \thead{$n$} & \thead{10th} & \thead{Q1 (25\%)} & \thead{Median} & \thead{Mean} & \thead{Q3 (75\%)} & \thead{90th} & \thead{IQR} \\
\midrule
\textbf{FrontierFinance} (ours)     & 220 & $-4.01$ & $0.19$  & $3.73$  & $\mathbf{4.84}$  & $10.56$ & $15.88$ & $\mathbf{10.38}$ \\
Finance Agent v2 (public)           &  27 & $-2.62$ & $-1.53$ & $-0.25$ & $0.50$  & $2.86$  & $3.70$  & $4.39$ \\
BigFinanceBench (public)            &  50 & $-4.18$ & $-2.57$ & $-1.26$ & $-0.35$ & $2.29$  & $6.32$  & $4.86$ \\
FinanceBench (\citep{islam2023financebench})   & 150 & $-8.31$ & $-6.96$ & $-5.10$ & $-5.42$ & $-3.75$ & $-2.61$ & $3.21$ \\
\bottomrule
\end{tabular}
\end{table}

Table~\ref{tab:positioning} reports where each benchmark's 
median query falls as a percentile of Samaya's \emph{entire} 
pool ($n = 4{,}212$ queries)
and how its queries split across difficulty buckets.
FrontierFinance's median query sits at the \textbf{80th percentile} of the internal
corpus while the external benchmarks
sit between the 16th and 49th percentile; correspondingly, a third of FrontierFinance
falls in the hard tercile versus 0--2\% for the externals.

\begin{table}[h]
\centering\small
\caption{Difficulty positioning on the shared BT scale (percentile vs.\ Samaya's internal distribution; higher = harder). Hard/Medium/Easy are FrontierFinance's own BT terciles.}\label{tab:positioning}
\begin{tabular}{@{}l r r r r r@{}}
\toprule
\thead{Benchmark} & \thead{$n$} & \thead{Median diff.\ pctile} & \thead{Hard} & \thead{Medium} & \thead{Easy} \\
\midrule
\textbf{FrontierFinance} (ours)   & 220 & \textbf{80.0} & 73 (33.2\%) & 74 (33.6\%) & 73 (33.2\%) \\
Finance Agent v2 (public)         &  27 & 48.6 &  0 (0.0\%)  & 11 (40.7\%) & 16 (59.3\%) \\
BigFinanceBench (public)          &  50 & 41.2 &  1 (2.0\%)  & 14 (28.0\%) & 35 (70.0\%) \\
FinanceBench (Islam et al., 2023) & 150 & 15.9 &  0 (0.0\%)  &  0 (0.0\%)  & 150 (100.0\%) \\
\bottomrule
\end{tabular}

\end{table}

\section{LLM API Endpoints Details}
\label{appendix:api-endpoint-details}
We used the Microsoft Azure OpenAI API endpoints for accessing the GPT series models. This also includes the GPT 5.5 Web Search harness, for which we used Azure OpenAI's built-in search-grounded API. We used the Google Vertex AI API endpoints for accessing the Gemini and Claude series models, as well as their Web Search harness versions. We used the Fireworks AI API endpoints for accessing the open-weight models, including the Kimi K3, GLM 5.2 and DeepSeek V4 Pro models.

\section{System and user prompts for the adapted Finance Agent v2 harness}
\label{appendix:finance-agent-prompt}

We include the system and user prompts of our re-implementation of the Finance Agent v2 harness below.
Compared to the original implementation, the only change is the added tool call limit paragraph in the middle of the system prompt.

\begin{promptbox}[System prompt]
You are a financial agent. You are given a question and you need to answer it using the tools provided. You will not be able to interact with the user or ask clarifications, you must answer the question only based on the information provided.

You should answer all questions as if the current date is {date}.

You will have access to a data storage system. You can use this system to store parsed contents of HTML pages retrieved from the web. You can then use the retrieve_information tool to apply answer questions or gather information from the stored documents using LLM-based prompts. This data storage system is designed to help you avoid context window issues.

When you have the final answer, you should call the `submit_final_result` tool with it. Your submission will not be processed unless you call this tool.

When making tool calls, you need to make sure you stay within the following limits:
- The maximum number of tool calls allowed (None for no limit): {max_tool_calls}
- The maximum wall-clock seconds allowed for tool calling (None for no limit): {max_tool_call_time}
If you are told that you have reached either of these limits, you must call the `submit_final_result` tool with the best answer you have gathered so far.
IMPORTANT: If any tool response returns with instructions indicating that tool call limit or time limit has reached, your VERY NEXT action MUST be to call `submit_final_result` with the best answer you have gathered so far. Do not make any other tool calls except `submit_final_result`. Failure to do so will cause the entire run to fail with no answer recorded.

You should include any necessary step-by-step reasoning, justification, calculations, or explanation in your answer. You will be evaluated both on the accuracy of the final answer, and the correctness of the supporting logic.

When possible, please provide any calculated answers to at least two decimal places (e.g. 18.78%

SEC filings are the most authoritative source of financial data. If a number appears in both an SEC filing and another source (e.g., a press release or company website), use the SEC filing's figure. You may freely use and cite non-SEC sources for information not available in SEC filings. For historical price data not available in SEC filings, use the `price_history` tool as your primary source. Fall back to `web_search` if the price tool is not working. You should always use the raw, unadjusted price data from the `price_history` tool, unless the question specifically asks for the adjusted price. Share prices should be reported in dollars with 2 decimal places, e.g. $10.25 per share. Stock indices (^IXIC, ^GSPC, ^SOX, etc.) are not covered by `price_history` - for index historical levels, start with an authoritative source such as the data provided by FRED. If the question references a specific source, make sure to incorporate information from that source, but still cross-reference SEC filings where relevant.

When reporting financial figures, use the same scale and units as presented in the SEC filing (e.g., if the filing reports values "in millions," report your answer in millions), unless otherwise specified in the question.

At the end of your answer, you should provide your sources in a dictionary with the following format:
{
    "sources": [
        {
            "url": "https://example.com",
            "name": "Name of the source"
        },
        ...
    ]
}
\end{promptbox}

\begin{promptbox}[User prompt]
Question:
{question}
\end{promptbox}

\section{System and user prompts for the grading LLM judge in \frontierfinance}
\label{appendix:llm-judge-prompts}

We include the system and user prompts used for the LLM judges below. The same prompts are used in our code release.

\begin{promptbox}[System prompt]
You are a senior financial analyst. Your task is to evaluate a financial report against a list of pre-defined rubrics. The report presented to you is generated to answer a specific financial query. For each given rubric, you are expected to produce a binary judgement on whether the rubric is satisfied or not by the financial report.

For each task, you will be given the following:
1. A financial query, which specifies the information the user is seeking.
2. The date the query was made. This is important for assessing the time understanding of the system. Whenever necessary, you should use this date as the temporal anchor for interpreting relative date terms in both the query and the rubrics.
3. A financial report which aims to answer that query.
4. One or more natural language rubrics, each checking a specific aspect of the report.

All of the input will be clearly marked in XML tags. Your task is to judge whether the report adequately satisfies each of the given rubrics. You must evaluate the report objectively and thoroughly.

Pay special attention to the following aspects when making your judgement:
1. **Each rubric should be judged independently**. Even in the case that one rubric seems related to another, you need to give your judgement of whether each rubric is satisfied independently.
2. **Pay attention to numerical units**. The report and the rubric might use different units to represent the same number. Take this into account when making your judgement. For example, "USD 2.1 billion" is equivalent to "USD 2,100 million".
3. **Accept reasonable numerical approximation**. A figure in the report is acceptable if it equals the rubric's figure after rounding the rubric's figure to the (coarser) precision the report uses. A figure stated at the same or finer precision than the rubric's, but with a different value, is NOT acceptable -- even if numerically close. For example, against a rubric value of "3,098 million": "3.1 billion" is acceptable (a correct rounding to two significant figures), but "3,105 million" is not (it asserts a precise, different value). Likewise against "7.14%
\end{promptbox}

\begin{promptbox}[User prompt]
You will evaluate the report below against the given set of rubrics. The report has been written to answer a specific query.

The query is provided below within the <query> tags.
<query>
{query}
</query>

The date the query was submitted is provided below within the <date> tags. This is important for assessing whether the report correctly understands the time aspect of the query.
<date>
{query_date}
</date>

The financial report is provided below within the <report> tags.
<report>
{report}
</report>

Now that you have read the query and the report, please evaluate whether the report satisfies each of the following rubrics. The list of rubrics is provided below within the <rubrics> tags. Each rubric is annotated with a unique ID, which you should use in your output to refer to that rubric.
<rubrics>
{rubrics}
</rubrics>

For each rubric, determine if the report adequately satisfies it. As a reminder, pay attention to the following aspects mentioned before:
- Each rubric should be judged independently.
- Pay attention to numerical units.
- Accept reasonable numerical approximation.

Your output must be ONLY a valid JSON object with the following structure:
```json
{
  "0": {
    "reason": "concise 1-sentence reason for your judgement on rubric 0",
    "label": true/false
  },
  "1": {
    "reason": "concise 1-sentence reason for your judgement on rubric 1",
    "label": true/false
  },
  ...
}
\end{promptbox}

\section{Tool call distribution for all models}
\label{appendix:toolsplit}

We provide the tool call distribution statistics of all models under the Finance Agent v2 harness in Table~\ref{tab:toolsplit}.

\begin{table}[h]
  \centering
  \small
  \caption{Share of each system's tool calls going to each tool available in the harness (\%). Calls per query represents the average number of tool calls over all answered queries. $R_{\text{all}}$ represents the macro-averaged qualification rate over all rubrics.}
  \label{tab:toolsplit}
  \begin{tabular}{@{}lcccccccc@{}}
    \toprule
    & \textbf{Web} & \textbf{Parse} & \textbf{Retrieve} & \textbf{EDGAR} & \textbf{Calcu-} & \textbf{Price} & \textbf{Calls /} & \textbf{$R_{\text{all}}$} \\
    \textbf{System} & \textbf{search} & \textbf{HTML} & \textbf{info.} & \textbf{search} & \textbf{lator} & \textbf{history} & \textbf{query} & \textbf{(\%)} \\
    \midrule
    Claude Fable 5   & 24.4 & 25.1 & 19.3 &  6.4 & 23.3 & 1.5 & 16.6 & 49.2 \\
    GPT 5.6 Sol      & 44.8 & 16.9 & 13.1 &  4.7 & 20.0 & 0.4 & 46.3 & 46.8 \\
    Kimi K3          & 36.5 & 26.6 & 18.7 &  6.8 &  9.9 & 1.5 & 19.6 & 46.4 \\
    Gemini 3.6 Flash & 31.1 & 20.0 & 21.8 & 11.0 & 15.4 & 0.7 & 24.8 & 46.3 \\
    Claude Opus 4.8  & 29.6 & 27.0 & 20.5 &  6.1 & 15.6 & 1.1 & 16.1 & 45.0 \\
    GPT 5.5          & 27.7 & 26.2 & 14.6 &  7.7 & 23.1 & 0.7 & 33.7 & 43.5 \\
    GLM 5.2          & 28.8 & 29.1 & 22.3 & 11.8 &  7.3 & 0.7 & 26.9 & 42.8 \\
    DeepSeek V4 Pro  & 45.2 & 21.4 & 18.5 &  4.5 &  9.9 & 0.5 & 41.1 & 40.5 \\
    Gemini 3.1 Pro   & 41.3 & 21.1 & 22.0 & 13.4 &  1.0 & 1.0 & 16.1 & 30.5 \\
    \bottomrule
  \end{tabular}
\end{table}

\section{Performance breakdown by use case}\label{sec:appendix_usecase_breakdown}

Table~\ref{tab:usecase-breakdown} reports the macro-averaged rubric qualification rate for each system broken down by use case.
Table~\ref{tab:rubric-breakdown} reports the micro-averaged rubric qualification rate for each system broken down by rubric categories.

\begin{table}[h]
\centering\footnotesize\setlength{\tabcolsep}{4pt}
\caption{Macro-averaged rubric qualification rate (\%) per use case for all evaluated systems, grouped by harness and ranked by $R_{\text{all}}$ within each group. Use case abbreviations: \textbf{Fin.}~=~Financial Data Extraction; \textbf{Earn.}~=~Earnings \& Events; \textbf{Co.}~=~Company Research; \textbf{Cov.}~=~Coverage \& Catalyst Monitoring; \textbf{Sec.}~=~Sector, Industry \& Macro; \textbf{Scr.}~=~Screening \& Discovery.}\label{tab:usecase-breakdown}
\begin{tabular}{@{}lc cccccc@{}}
\toprule
\textbf{System} & \textbf{$R_{\text{all}}$} & \textbf{Fin.} & \textbf{Earn.} & \textbf{Co.} & \textbf{Cov.} & \textbf{Sec.} & \textbf{Scr.} \\
\midrule
\multicolumn{8}{@{}l}{\textit{Web Search Harness}} \\
\quad Claude Opus 4.8   & 33.0 & 31.3 & 45.6 & 32.5 & 28.1 & 30.6 & 27.4 \\
\quad Gemini 3.1 Pro    & 30.7 & 29.1 & 41.3 & 29.7 & 32.1 & 25.0 & 27.2 \\
\quad GPT 5.5           & 20.7 & 17.3 & 26.1 & 24.5 & 27.0 & 14.0 & 21.7 \\
\addlinespace
\midrule
\multicolumn{8}{@{}l}{\textit{Finance Agent v2 Harness}} \\
\quad Claude Fable 5    & 49.2 & 55.6 & 63.9 & 40.5 & 49.0 & 38.1 & 33.3 \\
\quad GPT 5.6 Sol       & 46.8 & 48.9 & 58.0 & 42.1 & 46.7 & 40.9 & 36.5 \\
\quad Kimi K3           & 46.4 & 46.5 & 61.0 & 45.9 & 46.9 & 36.6 & 36.9 \\
\quad Gemini 3.6 Flash  & 46.3 & 56.2 & 54.0 & 41.2 & 36.5 & 36.4 & 36.1 \\
\quad Claude Opus 4.8   & 45.0 & 51.0 & 57.7 & 38.9 & 43.2 & 35.2 & 30.2 \\
\quad GPT 5.5           & 43.5 & 47.6 & 53.7 & 39.7 & 41.5 & 34.7 & 34.8 \\
\quad GLM 5.2           & 42.8 & 47.5 & 52.0 & 38.5 & 46.4 & 34.4 & 24.9 \\
\quad DeepSeek V4 Pro   & 40.5 & 45.3 & 54.5 & 32.1 & 38.5 & 32.7 & 28.1 \\
\quad Gemini 3.1 Pro    & 30.5 & 31.4 & 42.5 & 29.1 & 27.4 & 25.0 & 21.6 \\
\addlinespace
\midrule
\multicolumn{8}{@{}l}{\textit{Samaya In-house Harness}} \\
\quad Samaya (high effort) & 56.0 & 59.5 & 75.1 & 56.5 & 58.3 & 38.5 & 36.2 \\
\quad Samaya               & 52.9 & 55.3 & 71.0 & 52.8 & 57.8 & 38.7 & 28.6 \\
\bottomrule
\end{tabular}
\end{table}

\begin{table}[h]
\centering\footnotesize\setlength{\tabcolsep}{4pt}
\caption{Micro-averaged rubric qualification rate (\%) per rubric category for all evaluated systems, grouped by harness and ranked by $R_{\text{all}}$ within each group. $R_{\text{all}}$ is the overall macro-averaged qualification rate (as in Table~\ref{tab:usecase-breakdown}), included for system ranking purposes. Category columns are \textbf{micro}-averaged over all rubrics in each category. Category abbreviations: \textbf{Fact.}~=~Factual Data Extraction; \textbf{Qual.}~=~Qualitative \& Contextual Information; \textbf{Anly.}~=~Analysis \& Interpretation; \textbf{Comp.}~=~Comparative Analysis; \textbf{Fwd.}~=~Forward-Looking Information; \textbf{Fmt.}~=~Format \& Presentation.}\label{tab:rubric-breakdown}
\begin{tabular}{@{}lc cccccc@{}}
\toprule
\textbf{System} & \textbf{$R_{\text{all}}$} & \textbf{Fact.} & \textbf{Qual.} & \textbf{Anly.} & \textbf{Comp.} & \textbf{Fwd.} & \textbf{Fmt.} \\
\midrule
\multicolumn{8}{@{}l}{\textit{Web Search Harness}} \\
\quad Claude Opus 4.8   & 33.0 & 17.4 & 32.4 & 41.5 & 38.1 & 23.8 & 56.6 \\
\quad Gemini 3.1 Pro    & 30.7 & 20.0 & 25.6 & 41.9 & 42.9 & 25.9 & 53.4 \\
\quad GPT 5.5           & 20.7 & 16.0 & 29.9 & 40.2 & 39.5 & 24.6 & 59.3 \\
\addlinespace
\midrule
\multicolumn{8}{@{}l}{\textit{Finance Agent v2 Harness}} \\
\quad Claude Fable 5    & 49.2 & 43.1 & 41.2 & 49.9 & 50.2 & 48.0 & 74.4 \\
\quad GPT 5.6 Sol       & 46.8 & 35.7 & 44.2 & 52.5 & 53.0 & 43.4 & 79.4 \\
\quad Kimi K3           & 46.4 & 29.7 & 42.6 & 49.7 & 49.4 & 35.5 & 65.9 \\
\quad Gemini 3.6 Flash  & 46.3 & 38.0 & 36.5 & 48.6 & 45.5 & 39.2 & 74.1 \\
\quad Claude Opus 4.8   & 45.0 & 38.3 & 36.7 & 49.8 & 50.4 & 38.7 & 73.3 \\
\quad GPT 5.5           & 43.5 & 38.3 & 40.4 & 48.4 & 45.6 & 40.7 & 77.5 \\
\quad GLM 5.2           & 42.8 & 28.7 & 38.9 & 43.9 & 45.5 & 38.2 & 67.2 \\
\quad DeepSeek V4 Pro   & 40.5 & 30.2 & 36.5 & 44.3 & 39.8 & 29.6 & 65.1 \\
\quad Gemini 3.1 Pro    & 30.5 & 20.2 & 25.7 & 38.5 & 37.9 & 20.0 & 47.3 \\
\addlinespace
\midrule
\multicolumn{8}{@{}l}{\textit{Samaya In-house Harness}} \\
\quad Samaya (high effort) & 56.0 & 46.6 & 48.6 & 54.7 & 48.9 & 51.5 & 75.6 \\
\quad Samaya               & 52.9 & 40.7 & 47.1 & 55.7 & 52.3 & 46.0 & 69.5 \\
\bottomrule
\end{tabular}
\end{table}

\end{document}